\documentclass[11pt, letterpaper, twocolumn]{article}

\usepackage[english]{babel}
\usepackage[style=apa, natbib=true]{biblatex}
\usepackage[margin=0.5in]{geometry}
\usepackage[T1]{fontenc}
\usepackage[utf8]{inputenc}
\usepackage[multiple]{footmisc}
\usepackage[dvipsnames]{xcolor}
\usepackage{amsmath}
\usepackage{amsfonts}
\usepackage{bm}
\usepackage{graphicx}
\usepackage{graphbox}
\usepackage{csquotes}
\usepackage{setspace}
\usepackage{caption}
\usepackage{subcaption}
\usepackage{wrapfig}
\usepackage{tabularx}
\usepackage{enumitem}
\usepackage{multirow}
\usepackage{siunitx}
\usepackage{authblk}
\usepackage{url}
\usepackage{hyperref}

\newcommand{\bb}[1]{\mathbb{#1}}
\newcommand{\cl}[1]{\mathcal{#1}}
\newcommand{\tbf}[1]{\textbf{#1}}
\DeclareMathOperator{\argmin}{arg\,min}

\AtBeginBibliography{\footnotesize}
\AtEveryCite{\color{MidnightBlue}}

\title{Transferable Class-Modelling for Decentralized\\Source Attribution of GAN-Generated Images}
\author[*]{Brandon B. G. Khoo}
\author[*]{Chern Hong Lim}
\author[**]{Rapha\"{e}l C.-W. Phan}
\affil[*]{School of Information Technology, Monash University Malaysia}
\affil[**]{Faculty of Information Technology, Monash University}
\date{March 2022}

\begin{document}
\maketitle
\setlength{\parskip}{0.5em}

\begin{abstract}
    GAN-generated \textit{deepfakes} as a genre of digital images are gaining ground as both catalysts of artistic expression and malicious forms of deception, therefore demanding systems to enforce and accredit their ethical use. Existing techniques for the source attribution of synthetic images identify subtle intrinsic fingerprints using multiclass classification neural nets limited in functionality and scalability. Hence, we redefine the deepfake detection and source attribution problems as a series of related binary classification tasks. We leverage transfer learning to rapidly adapt forgery detection networks for multiple independent attribution problems, by proposing a semi-decentralized modular design to solve them simultaneously and efficiently. Class activation mapping is also demonstrated as an effective means of feature localization for model interpretation. Our models are determined via experimentation to be competitive with current benchmarks, and capable of decent performance on human portraits in ideal conditions. Decentralized fingerprint-based attribution is found to retain validity in the presence of novel sources, but is more susceptible to type II errors that intensify with image perturbations and attributive uncertainty. We describe both our conceptual framework and model prototypes for further enhancement when investigating the technical limits of reactive deepfake attribution.
\end{abstract}

\section{Introduction}

\begin{figure*}[ht]
    \centering
    \includegraphics[width=\linewidth]{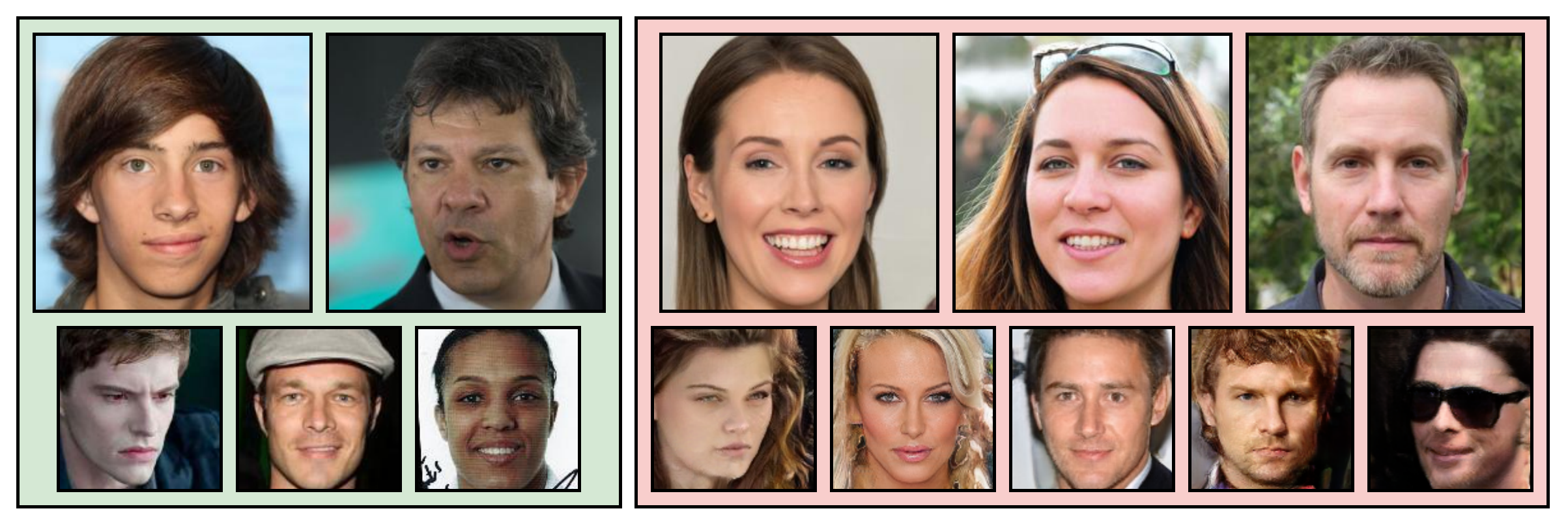}
    \caption[Examples of real and synthetic images]{Examples of real human portraits (left) and synthetic images (right) from different sources in the datasets we used. Can you spot the differences with the naked eye?}
    \label{fig:image_collage}
\end{figure*}

Consider the following scenario straight out of a cyberpunk novel: exceptionally photo-realistic images created with deep neural networks (Figure \ref{fig:image_collage}) dominating the consumption of digital media and obliterating virtually any fighting chance of distinction between fact and fiction. These images exist today under the umbrella terms of \emph{deepfakes} and \emph{synthetic media}, often via on-demand content generation services, and applied in contexts both constructive and destructive toward society \citep{Chesney_2018}. To date, statistically synthesized fictional photographs and human avatars are revolutionizing the communications and entertainment industries, while also subjecting societies to menacing new forms of disinformation and fraud. Ad-hoc methods have been devised for synthetic image recognition beyond human capacities, though they are engaged in a technical arms race against the \textbf{generative models\footnote{Deep neural networks designed to create, modify, or predict instances of data. Examples include GANs and VAEs.} (GMs)} producing these images. This unsustainable situation is exacerbated by the advent of highly proficient GMs and their mass adoption on social media \citep{Westerlund_2019}, thus leading to new security challenges and philosophical questions alike.

In digital image forensics, three attributes are commonly inquired about any image: its authenticity (validity), provenance (origin), and integrity (presence of anomalies). Knowledge of these attributes safeguards the dissemination and interpretation of digital images. However, contemporary GMs like StyleGAN \citep{Karras_2019} and its derivatives can conjure high-definition synthetic images of people, objects, and scenes, disrupting the utility of established forensic methods that assume the use of cameras, conventional CGI, or image editing software. Various data-driven \textbf{deepfake detection models} currently exist to detect GM forgeries, but their obscure and dichotomous approaches are often inappropriate and inconclusive in practice \citep{Lyu_2020}. These models commonly fail to generalize beyond conditions expected during training \citep{verdoliva2020media}, including novel sources and image post-processing. Reliance on deepfake detection alone has a fundamental flaw in that it implicitly assumes all deepfakes are made with malicious intent; that there is value in thoroughly separating them from ``real'' content despite the nuances imposed by the increasing adoption and heterogeneity of deep generative modelling.

Aside from shifting definitions of image authenticity, there are two real-world scenarios where deepfake detection is practically insufficient \citep{yu2020artificial, kim2020decentralized}: intellectual property infringements and irresponsible GM use. The former is a natural consequence of burdensome resource costs for optimising GMs, which may drive unscrupulous individuals towards plagiarism or unauthorized use of pretrained GMs; thereby stifling innovation and profiting off others' investments. This demands solutions to reliably \emph{attribute} synthetic images to their rightful proprietors despite provenance masking attempts. The latter scenario concerns the \emph{dual-use} potential of image generation by both developers and end users, where the involvement of GMs alone does not imply harmful intent deserving of intervention. In this situation, reliance on deepfake detection risks aggravating the \emph{liar's dividend}\footnote{The failure to uphold non-repudiation of digital content due to the proliferation of untraceable GMs. For instance, a criminal caught on camera can deny any wrongdoing by invoking the plausibility of perfectly fabricated evidence.} if no meaningful justifications or supporting materials are provided to refute deepfake claims. \emph{Source attribution} of synthetic images therefore offers a partial solution to alleviate both issues by leveraging the media literacy and judgment of users, providing evidence to assist them in forming their own contextually grounded conclusions.

We propose a semi-decentralized classification framework to ascertain the involvement of specific GMs in images that can be encountered online. Our neural network-based source attribution method determines whether images contain characteristic traces of GM manipulation, and if a particular GM class can be held responsible for them. Individual fake (synthetic) images from a common origin are identified and distinguished from both real images presumably acquired via traditional means, as well as fake images from all other known sources in the same domain. Complete models based on this framework leverage \emph{transfer learning} of common source-agnostic features derived from the image forgery detection task (hence, ``transferable'') to accelerate specialized training for source attribution tasks on different GMs, exploiting the conceptual overlap of both forensic scenarios for added developmental efficiency.

Like previous attempts at deepfake source attribution \citep{Yu_2019, Frank_2020}, our method relies on supervised learning of fingerprint-like signatures intrinsically incorporated within all synthetic images to date, and thus susceptible to all its caveats and theorized constraints. However, we adopt a modular approach to attribution whereby the practical advantages of image forgery detection can be retained. By only requiring the fitting of inclusive decision boundaries to solve a binary classification task rather than an exhaustive categorical one, our method is technically capable of handling situations such as indeterminate or coincident GM fingerprints. Although it is obviously impossible to ascribe images to novel sources, their manifested fingerprints should fall within the absolute complement of known sources barring exceptionally novel cases. Moreover, we disentangle source attribution from deepfake detection without necessitating entire distinct classifiers, thereby increasing scalability, extensibility, and ease of maintenance.

In our proof of concept experiments, we investigate whether synthetic image source attribution can be practically modelled as a binary classification problem in an open-set context; whereby GMs are multitudinous yet often reused and hence demand measures for intellectual property protection. We also assess the strengths and deficiencies of our proposed model implementation against similar baselines within their pre-established methodologies, and analyse the additional challenges faced when dealing with degraded images. Our results affirm the relative ease of deepfake detection on most lower-resolution images of human faces, but cast unsettling doubts on the long-term sustainability of deepfake attribution without the aid of steganographic additives.

The key contributions of this work are threefold: Firstly, we apply the more adaptable one-versus-all binary classification strategy to fingerprint-based reactive attribution of synthetic image sources. Secondly, we conjecture that source attribution is a specialized case of deepfake detection, and develop an extensible classification framework exploiting that assumption with transfer learning to ameliorate the inefficiencies of one-versus-all classification. Thirdly, we demonstrate the potential benefits and consequences of implementing the aforementioned design decisions, including descriptive model interpretation via class activation mapping.

\section{Related Work}

Generative adversarial networks (GANs) \citep{goodfellow2014generative} are often the preferred deep generative models (GMs) for synthetically creating or manipulating realistic images, due to their superior adaptability and output quality despite the notorious difficulty of optimising them. The success of the GAN framework spearheaded by the likes of StyleGAN, ProGAN \citep{karras2018progressive}, and BigGAN \citep{brock2018large} among others have since solidly associated adversarial learning techniques with synthetic media. Conditional GANs have since been applied to enhance CGI cinematography, streamline teleconferences, and produce deepfakes of varying ethical acceptability. The latter has provoked the introduction of image forgery detection systems to identify characteristic features of GAN-generated images where human eyes fail to notice, often making use of deep learning to take advantage of subtle statistical cues \citep{Hsu_2018, Hsu_2020, Marra_2018, Nataraj_2019, Zhang_2019, Wang_2020_FakeSpotter, Durall_2020}. Determined to future-proof their forensic detectors against the continual emergence and rapid development of novel GMs, some researchers emphasized adaptability measures including few-shot learning strategies \citep{cozzolino2018forensictransfer}, extraction of common discriminative features \citep{Hsu_2020}, and data augmentation to simulate non-ideal conditions \citep{Wang_2020_ForenSynths}.

Concurrently, source camera attribution is playing an increasingly instrumental role in digital image forensics. Many relevant works in that regard analyse the residual noise signals obtained by subtracting from naturally sourced images their denoised, purely semantic versions. \citet{lukavs2006detecting} famously shown that the noise residuals of all images captured by the same device contain the same Photo Response Non-uniformity (PRNU) patterns resulting from manufacturing imperfections, which draw comparison with human biometric \emph{fingerprints}. \citet{cozzolino2019noiseprint} then proposed a Siamese convolutional network for extracting ``noiseprints'' from images where model-based artefacts are amplified. By training the network to minimize the distance between \emph{patches} of noise residuals from images captured by the same camera model, while maximizing the distance between patches associated with different camera models, the resulting noiseprints clearly indicate anomalous image regions and provide hints for model-level attribution. Despite the artificial nature of computer-generated images that lack camera-related impressions \citep{Böhme_2012}, it is now known that GAN-generated images exhibit their own characteristic artefacts analogous to PRNU patterns \citep{Marra_2019}, which led some to question if all images generated using convolutional neural networks (ConvNets) inherently contain distinctive traces that dissociate them from both real images and images by other ConvNets, and to what extent could existing forensic practices be applicable to them.

\subsection{Reactive Attribution} \label{reactive}

Synthetic image source attribution has since broken ground with data-driven classifiers capable of near-perfect accuracy in ideal conditions, albeit assuming finite sets of plausible source GMs consisting largely of somewhat obsolete GANs. \citet{Yu_2019} initiated a ConvNet-based attribution method that detects GAN-generated images and predicts either the architectural model or exact instance of the GM that created them. Their ``attribution network'' learns internal representations of supposedly unique GM fingerprints (functionally identical to noise residuals) originating from the training parameters of different GMs, and compares them with fingerprints encoded from the image probed at runtime to predict its source GM; though the desired level of attribution varies depending on the dataset. \citet{Frank_2020} later improved on the fingerprint-based attribution method while substantially simplifying the attribution network's complexity. They utilized up-convolutional signatures in the frequency domain, where fingerprints may manifest as prominent periodic patterns when averaged over many samples \citep{Zhang_2019, Wang_2020_ForenSynths}. Both studies adopted multiclass recognition for simultaneous authentication and attribution, which renders them incapable of properly processing blatantly fake images derived from unknown sources, among other limitations.

Another successful GM attribution experiment was done by \citet{goebel2020detection} on six prevalent GANs, using RGB co-occurrence matrices as features computed from the probed images \citep{Nataraj_2019} in lieu of fingerprint recognition. Their method could also achieve visual \emph{interpretability} via feature localization: image regions of synthetic origin are predicted by processing and classifying each image as individual patches, which are then presented as saliency maps that help describe the underlying reasoning in a user-friendly manner. More recently, \citet{ModelParsing2021} introduced a radically different approach to GM attribution dubbed ``model parsing''. Their dual-classifier model indirectly determines source GMs down to their discretized hyperparameters, based on optimal fingerprints estimated from probed images. While their method could predict the architectural details of most GMs even if excluded from training data, its practical applications might be limited by its incomplete attribution whereby exact sources must be deduced from said predicted details.

\subsection{Proactive Attribution} \label{proactive}

Aside from \cite{ModelParsing2021}, all other attempts mentioned thus far are supervised learning methods trained on predefined image-source pairs $(I,y), I \sim \bb{I}, y \in \bb{Y}$; where $\bb{I}$ is composed of both real and fake images, and $\bb{Y}$ is the set of sources (cameras or GMs) being considered. Although suitable as baselines for further research, they are inherently inscalable, rapidly obsolescent, and uncertain in reality where plausible sources are not bounded to the arbitrarily learned set of GMs. Consequently, proactive methods involving the embedding of steganographic keys or watermarks during content generation were proposed to facilitate attribution afterward with theoretical assurance. \citet{kim2020decentralized} devised an alternative paradigm of \emph{decentralized attribution} using different binary classifiers per GM, parameterized by keys directing the perturbations of output images. Optimal conditioning of the keys ensures the \emph{distinguishability} of generated images from real ones, and the \emph{attributability} of generated images of differing provenance from each other, without excessively compromising image quality. 

Meanwhile, \citet{yu2020artificial} developed a black-box method of watermarking any GM to defend it against plagiarism/abuse, by embedding watermarks into the training dataset of real reference images and then recovering said watermarks from generated images using jointly trained encoders and decoders respectively. Both works sought to overcome the perceived limitations of intrinsic GM fingerprints by placing the burden of attribution on watermark identification, which has unavoidable (albeit generally imperceptible) negative effects on GM outputs, and depends on the compliance of GM developers themselves towards established watermarking guidelines. An irresponsible developer could train their GM on non-watermarked data, whether intentionally or otherwise, thus evading the attributive infrastructure.

\subsection{Practical Challenges} \label{challenges}

Regardless of how image source attribution is performed, all methods centred on recognizing signal-level features are vulnerable to image processing operations that dilute, conceal, or even replace critical features. This ranges from perturbations as a result of innocent post-processing to ease online transmission (a phenomenon dubbed ``laundering'' \citep{Lyu_2020} when maliciously exploited), to more advanced attacks such as adversarial examples \citep{Carlini_2020}, fingerprint spoofing \citep{cozzolino2019spoc}, and fingerprint removal \citep{Neves_2020}. Of these challenges, the most commonly discussed is the lossy compression of JPEG images\footnote{Unlike simple perturbations, JPEG compression can be implemented differently due to its many degrees of freedom, limiting the utility of experiments applying it.}, which is often claimed to have destructive effects on both the noise residual fingerprints and frequency domain artefacts of synthetic images. \citet{mandelli2020training} also highlighted the relevance of the 8x8 pixel grid imposed on JPEG images during the Discrete Cosine Transform (DCT) block compression process, by demonstrating the failure of detection and attribution networks when evaluated on randomly cropped images. The pre-established approach to mitigate the effects of image post-processing (e.g. blurring, resizing, cropping, rotation, relighting, and additive noise) is to retrain the classifiers on samples that include different configurations and/or combinations of these perturbations, thereby (partially) resolving the discrepancy between training and testing data and forcing the networks to emphasize perturbation-resistant mesoscopic features if possible \citep{Yu_2019, Wang_2020_ForenSynths, Frank_2020}. However, to the best of our knowledge, no comparable remedy exists for the more alarming adversarial attack scenarios.

\section{Design and Methodology}

\subsection{Proposed Framework} \label{framework}

\begin{figure*}[ht]
    \centering
    \includegraphics[width=0.9\linewidth]{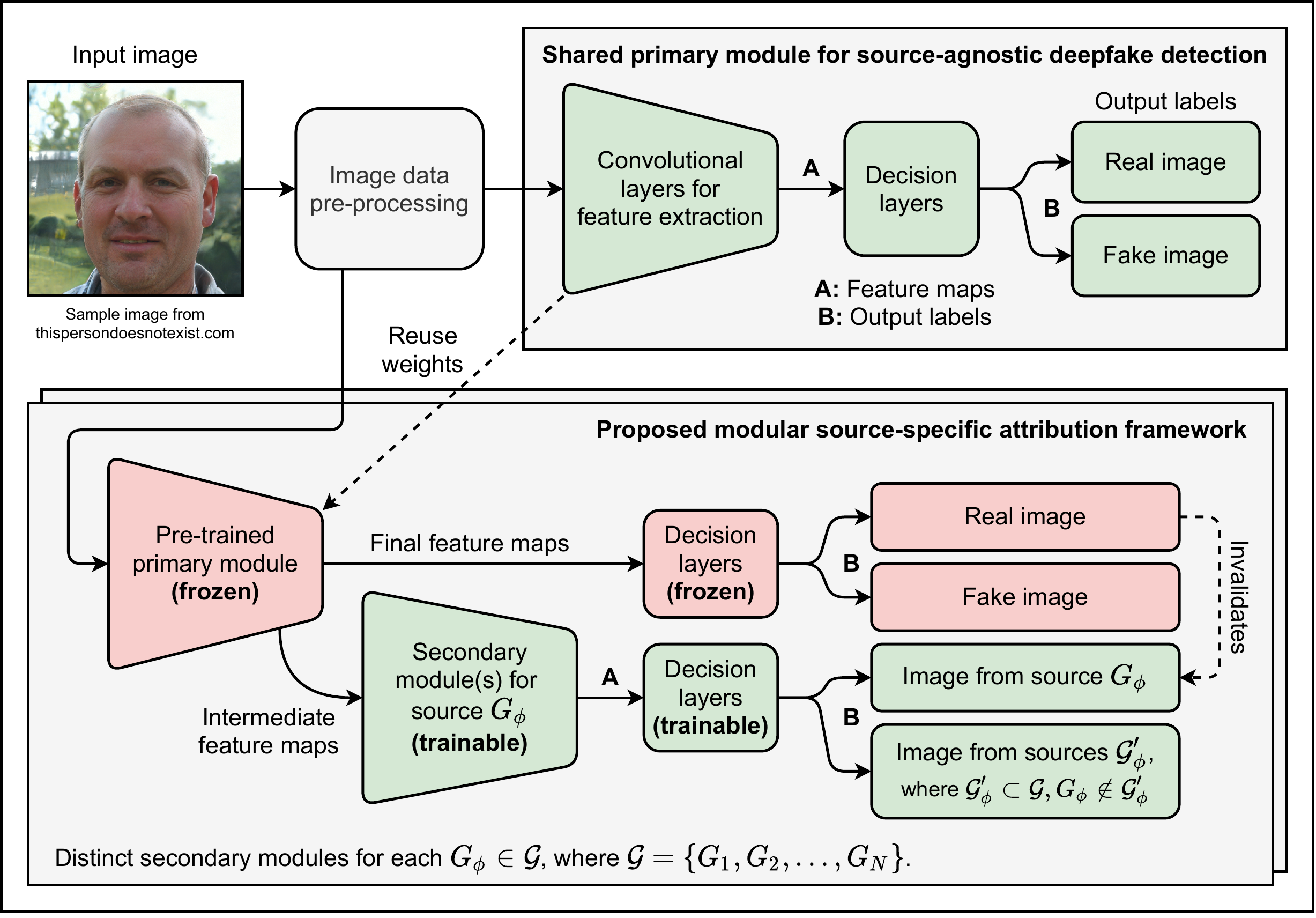}
    \caption[High-level diagram]{High-level diagram of the proposed framework for the decentralized source attribution of GAN-generated images. Different instances of the classifier (bottom) are trained to differentiate real images from synthetic images of different sources $G_{\phi} \in \cl{G}$, with all classifiers sharing common convolutional layers (top) initially trained for image forgery detection.}
    \label{fig:generator_attribution}
\end{figure*}

To address the technical weaknesses of previous methods for reactive attribution of synthetic images mentioned in section \ref{reactive}, we propose a modular semi-decentralized image recognition framework that disentangles the deepfake detection and source attribution tasks without incurring excessive computational overhead. Each classification model based to our design is composed of at least two modules of convolutional neural layers with individually distinct outputs. The \textbf{primary module} is optimized for source-agnostic deepfake detection, but also mostly functions as a pre-trained feature extractor for any number of \textbf{secondary modules}, each tasked with identifying traces of one specific generative algorithm in relation to all other plausible sources defined during training. We argue that framing source attribution as a series of binary classification problems would help resolve uncertainty in situations involving synthetic images generated by hitherto unknown or ambiguous sources. Additionally, according to the expected demands of commercialized deepfake generation services, the desire to determine culpability can be abridged to merely establish whether a certain GM implementation was involved in image creation; knowledge of other sources being viewed as less important.

Rather than maximizing statistical distances between different learnable GM fingerprints, the classification task at the attribution level is simplified to only recognize the presence of specific fingerprints. We intend to determine if this alternative approach, theoretically verified for watermark-based proactive attribution \citep{kim2020decentralized}, remains feasible when limited to intrinsic, naturally occurring GM fingerprints. This is in line with current research priorities about determining the limits of what information can be garnered from such fingerprints. The hypothesized downsides of relying on fingerprint recognition include heightened classifier vulnerability to ``laundering'' attacks and loss of guarantees on distinguishability and attributability, which can be compensated (to an extent) with larger, augmented training datasets. Since distinguishability from the real is necessary but insufficient for attributability among other fakes, the desired solution must fulfill both conditions albeit in decreasing order of importance:
\begin{enumerate}
    \item Detect if an image $I$ lies within the distribution of real images $P_{\cl{D}}$ or synthetic images $P_{\cl{G}}$.
    \item Predict the source GM, $G_{\phi} \in \cl{G}$ of image $I$, where $\cl{G} = \{G_1, G_2, \dots, G_N\}$, if $I$ is fake.
\end{enumerate}

The value of $N$ is unattainable in an open-world scenario as new image sources constantly emerge. To address this constraint, individually trained binary classifiers $F_{\phi}(I)$ are each associated with a specific source GM of interest $G_{\phi}$, to distinguish between classes $G_{\phi}$ and $\cl{G}_{\phi}^{\prime}$ where $\cl{G}_{\phi}^{\prime} \subset \cl{G}, G_{\phi} \not\in \cl{G}_{\phi}^{\prime}$. Assuming that ideal fingerprints that maximize attributability can be learnt for any designated $G_{\phi}$ , the classifier should satisfy the requirements for one-versus-all attribution regardless of the GM developers' support. An obvious problem with this approach is that different classifiers $F_{\phi}(I)$ must be procured for all existing $G_{\phi} \in \cl{G}$, which is impractical if every classifier is trained from scratch. However, since deepfake detection is necessary for synthetic image attribution and considering that the latter is a specialization of the former, the successful recognition and identification of any fingerprint or watermark known to indicate manipulation (barring adversarial examples) is sufficient to doubt the veracity of any given image prior to further analysis. Computational redundancies can therefore be reduced by attaching low-level feature extraction layers from the common primary module to specialized high-level layers that maximize attributability. Even without fine-tuning, the adoption of transfer learning enables a robust and adaptable deepfake detection model to be rapidly extended for various attribution tasks where demanded, as described in Figure \ref{fig:generator_attribution}.

\subsection{Model Implementation} \label{models}

\begin{figure}[!t]
    \centering
    \includegraphics[width=\linewidth]{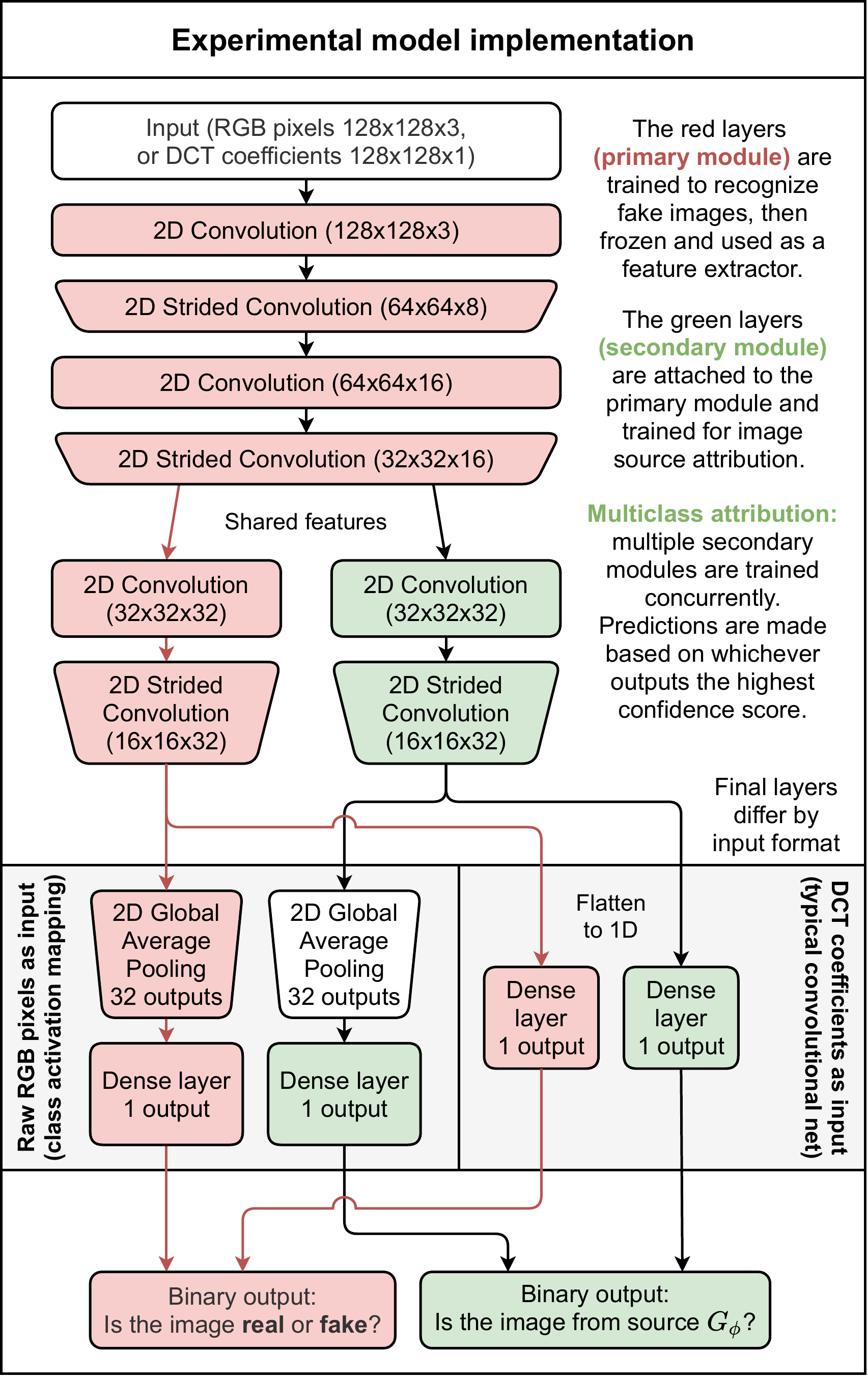}
    \caption[Proposed model diagram]{Topological diagram of our implementation of the classification framework specified in Figure \ref{fig:generator_attribution}. The annotations assume 128x128 image inputs used in the GANFP dataset; all convolutional tensors have their spatial resolutions doubled when trained on 256x256 FacesHQ+ images. The proposed model may consist of any number of secondary modules trainable for image source attribution, all obtaining feature maps from a common pre-trained primary module.}
    \label{fig:model_diagram}
\end{figure}

We now describe our proof-of-concept implementation of the proposed framework as detailed in Figure \ref{fig:model_diagram}. The primary module has a conspicuously shallow and simplistic structure that is sufficient for our experiments, but can theoretically be substituted with more complex classifiers. Double strided convolutions for downsampling are included every other layer in similar fashion to the PatchGAN discriminator topology \citep{Isola_2017}; since they have the advantage of retaining more relevant fine-scale information compared to intermediate pooling operations. Batch normalization is also applied in all but the first convolutional layer for improved generalization and rapid model fitting. The secondary modules used for source attribution tasks branch out after four layers and replicate the primary module structure for the remaining layers. When training individual secondary modules, common feature map inputs are obtained as intermediate outputs from a frozen, pre-trained primary module.

The output layer of each module is preceded by \emph{global average pooling} (GAP), whereby the feature maps processed through all previous layers are summarized and weighted in a \emph{class activation mapping} (CAM) structure \citep{Zhou_2016} to facilitate visual localization of image regions containing relevant features leading to positive classifications. We drew inspiration from the ``post-pooling'' variant of \citet{Yu_2019}'s attribution model, which consists of several successive average pooling operations prior to the final convolutional block. Since GAP is inappropriate for summarizing spectral coefficients, variants of our model intended for frequency domain analysis omit CAM for a typical ConvNet structure at the cost of increased complexity. Grad-CAM \citep{Grad-CAM} is applied instead on these models.

For multiclass attribution, multiple independent secondary modules $A_G = \{A_1, A_2, \dots, A_N\}$ are individually optimized for attributing specific GMs $\cl{G} = \{G_1, G_2, \dots, G_N\}$, with the final prediction $G_{\phi}$ obtained by taking the maximum value across all secondary sigmoid outputs according to the function: 
\[ G_{\phi} = \argmin_{G \in \cl{G}} A_{G}(I) \]
The output of each secondary module may not be mutually exclusive with respect to others built upon the same primary module. It is technically possible for an image to be predicted \texttt{fake} without being successfully ascribed to any source, assuming that the primary module has learnt fingerprint elements generalizable beyond the training distribution. Likewise, it is also possible for our model's secondary output(s) to \emph{contradict} its own primary output (e.g. predicting an image as \texttt{real} with at least one positive attribution). We recorded the \textbf{self-contradiction rate (CR)} in our experiments, but ideally the primary output (deepfake detection) should be prioritized during any such occurrence.

\begin{figure}[!t]
    \centering
    \includegraphics[width=\linewidth]{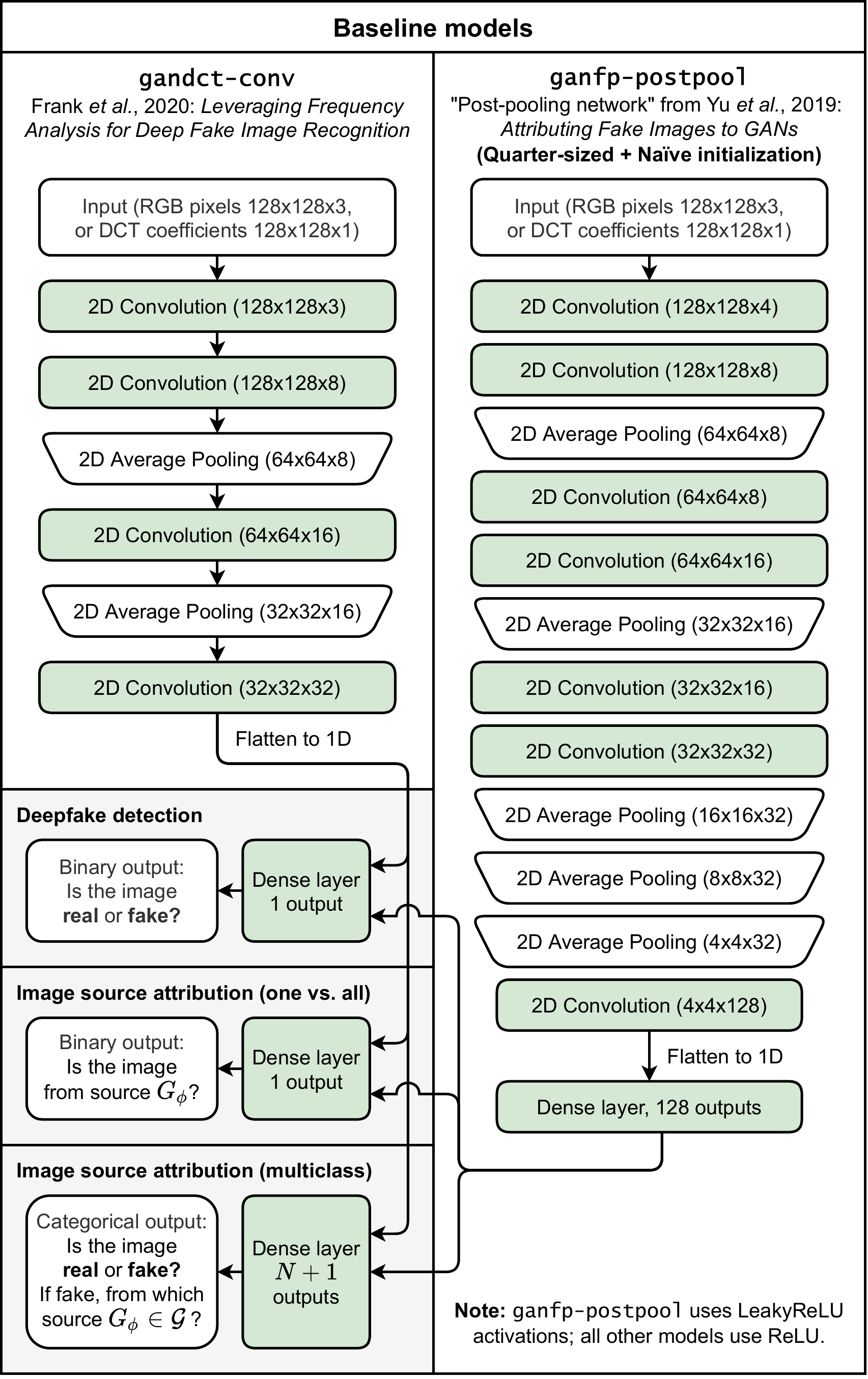}
    \caption[Baseline model diagrams]{Topological diagrams of the baseline models adopted from \citet{Frank_2020} and \citet{Yu_2019} respectively. Different decision layers are implemented depending on the scenario. The \texttt{ganfp-postpool} model evaluated in our experiments uses LeakyReLU activations, and contain only 1/4 of the original's feature maps due to resource constraints.}
    \label{fig:baselines_diagram}
\end{figure}

To benchmark our proposed model against previous studies, we adopt as performance baselines two ConvNet models designed for deepfake source attribution: the minimalist \texttt{gandct-conv} used by \citet{Frank_2020} in their frequency domain experiments, and a scaled-down version of \citet{Yu_2019}'s post-pooling attribution model \texttt{ganfp-postpool}. The rationale for selecting these models concerns their similarities and compatibility with our model. While the \texttt{gandct-conv} has a similar but slightly increased number of trainable parameters compared to our model, the \texttt{ganfp-postpool} contains three times as many parameters\footnote{The full-size \texttt{ganfp} model contains up to \num{9e6} parameters.}. Ideally, model complexity for forensic tasks should be constrained to increase developmental efficiencies and avoid risks including overfitting and adversarial vulnerabilities.

Figure \ref{fig:baselines_diagram} describes the structure of the baseline models, including the different decision layers applied based on whether binary (sigmoid) or multiclass (softmax) classification is being performed. In contrast with our model implementation, both baseline models lack batch normalization and use average pooling operations for downsampling. These models can be purposed for either deepfake detection or source attribution. However, each individually trained instance of these models can only perform either of them exclusively, since the entire model is optimized for a single purpose without support for multi-label processing. Without interchangeable feature extractors administering common inputs, we must retrain the baseline models entirely from scratch when switching between attribution tasks, impeding development. 

To the best of our knowledge, there are no previous studies on reactive deepfake attribution as a binary classification problem, so we are also investigating the performance of the baseline models when repurposed for this new scenario. It should be noted that unlike our proposed model, the baseline models are technically incapable of distinguishing between a failed attribution and successful recognition of a fake image from an unrelated source; neither are they equipped for anything other than a 1:1 correspondence between fake images and GMs. Additionally, the baseline models can only process images of the same spatial resolution as their initially fitted dataset once instantiated, whereas the fully convolutional CAM structure of our model (raw pixel input variant) enables larger images to be directly processed without amplifying model complexity. As such, reasonable comparisons are possible only on lower-resolution datasets.

\subsection{Datasets and Augmentations} \label{datasets}

To reduce the chance that predictions made by the proposed classifier will eventually be invalidated by the introduction of novel generative algorithms, a diverse dataset of relevant images is required to establish sturdy class-modelling boundaries between the feature distributions of considered sources. This includes definitive plenty of real images of different origin, fake image samples for each source GM $G_{\phi}$, and some held-out images from external sources $\cl{G}_{\phi}^{\prime}$ for use as negative observations. For our experiments, a miniature dataset of 52,000 human portraits termed \textbf{FacesHQ+} was derived from the existing FacesHQ dataset \citep{Durall_2020} containing 10,000 images/source for two sources of real photographs (CelebA-HQ\footnote{The CelebA-HQ dataset was actually upsampled from CelebA \citep{CelebA} using a GAN-based technique, so its categorization as a \emph{real} image set is disputable.}, FFHQ) and two sources of StyleGAN-generated portraits, namely \texttt{ThisPersonDoesNotExist.com} (TPDNE) and the \texttt{generated.photos} project. We extend FacesHQ with 10,000 additional images from the StyleGAN2 \citep{Karras_2020} version of TPDNE trained on FFHQ, and a set of 2,000 images created by StarGANv2 \citep{Choi_2020} from CelebA-HQ templates\footnote{StarGANv2 functions as a conditional GM for style transfer, combining stylistic elements of images to produce new ones.} to serve as out-of-distribution \emph{external samples} intentionally excluded from the training set to evaluate external validity. We adopt a train/validation/test ratio of 7:1:2 for samples from each source. The adaptation of our experiments to image domains other than human portraits (e.g. LSUN dataset) is left for future work.

Additionally, for benchmarking purposes we utilize the \textbf{GAN Fingerprints} (GANFP) dataset, which was also applied in \citep{Yu_2019, Frank_2020, ModelParsing2021}, albeit now truncated to 20\% of the original size due to resource constraints. We reuse the same GAN instances pre-trained by \citet{Yu_2019} to generate images of resolution 128x128, and specify a train/validation/test ratio of 67:13:20. This 150,000 image dataset consists of 30,000 PNG images for each of 5 source classes: real, SNGAN \citep{SNGAN}, ProGAN \citep{karras2018progressive}, MMDGAN \citep{MMDGAN}, and CramerGAN \citep{CramerGAN}. Note that the latter two GAN sources are implemented mostly identically, which is expected to result in somewhat similar fingerprints. CelebA \citep{CelebA}, which belongs in the same domain as FacesHQ+, is adopted as the sole source of real (albeit erratic) images. In this case, both the real and GAN-generated images are of mediocre visual quality prior to augmentation; and like higher-resolution FacesHQ+ images, occasionally contain salient high-level artefacts (obvious to the naked eye). The ratio of real to synthetic images is 1:4 in the GANFP dataset, compared to 2:3 in FacesHQ+. Since all real-labelled images have undergone some degree of homogeneous post-processing, we acknowledge that this might negatively affect the models' behaviour if real images are unintentionally associated with a specific signature pattern, rather than by the absence of it.

All images are initially downsampled to square images of 128x128 (for GANFP) or 256x256 (for FacesHQ+) spatial resolution, which reduces the resource requirements of the classifier and appropriately simulates the mainstream contemporary use of GAN-generated portraits as easily transmissible thumbnails and social media avatars. However, this compromises the presence and scale of discriminative features (and may also introduce new confounding features during rescaling), thus increasing the risk of incorrect predictions. Augmented versions of the dataset are also prepared to evaluate classifier robustness. Following established practice \citet{Frank_2020}, we apply each of the following post-processing perturbations to the entire dataset:
\begin{enumerate}
    \item Gaussian \textbf{blurring} with kernel size randomly sampled from $(3,5,7,9)$. This is known to destroy high-frequency fingerprints.
    \item Random \textbf{cropping} along both axes, with the percentage to crop uniformly sampled from $U(5,20)$. This disrupts the 8x8 JPEG grid. Cropped images are upsampled to the resolution of the originals.
    \item JPEG \textbf{compression} with quality factor (QF) uniformly sampled from $U(10,75)$. Subtler effects compared to other augmentations.
    \item Additive i.i.d. Gaussian \textbf{noise} with variance uniformly sampled from $U(5,20)$. This may add fingerprints where there should be none.
\end{enumerate}
In the multi-augmentation case, approximately 92-94\% of the dataset is subjected to at least one of the aforementioned augmentations, each applied with 50\% probability but always in that specific order. Considering that intensely degraded images are relatively unlikely in the wild, we also prepare separate copies of the dataset whereby individual augmentations (either JPEG compression or random cropping) are applied to half of the images within.

\subsection{Experimental Setup} \label{setup}

All our experiments were performed on a desktop PC equipped with a single nVidia GeForce RTX 2080 GPU with 8GB VRAM. We use Keras and TensorFlow v2 as our deep learning API and framework respectively, except for legacy code (e.g. GANFP GMs) which necessitate the use of TensorFlow v1. Our experimental random seeds are \texttt{0} for data generation and pre-processing, and both \texttt{2021} and \texttt{1000} for classifier model training and evaluation. We use na\"{i}ve Gaussian weight initialization with standard deviation 0.02 when training new model instances from scratch. All training phases are limited to at most 100 epochs, with early stopping regularization conditioned on cross-entropy loss on the validation set. We use the Adam optimizer, set to learning rates of \num{1e-3} for deepfake detection and \num{1e-4} for source attribution\footnote{Initial experiments have shown the models to fail at converging in attribution scenarios with learning rate \num{1e-3}.}. Minibatch sizes 256 and 128 are used for GANFP and FacesHQ+ datasets respectively. All baseline models are evaluated only on the GANFP dataset.

We develop variants of each classifier model trained on images in two different representations: the \emph{spatial domain} (raw RGB image \textbf{pixels}) and \emph{frequency domain} (\textbf{DCT} spectra obtained via type-II 2D-DCT, also used in JPEG compression), each with their own signal-level features. According to \citet{Zhang_2019} and \citet{Wang_2020_ForenSynths}, the artefacts associated with synthetic images are mainly concentrated in the mid-high frequency components, but also exist across the spectrum. Images represented in the frequency domain are first converted to greyscale before applying DCT, followed by log-scaling and normalization based on the statistics of the clean training set. Frequency domain input is not applied for FacesHQ+ due to the limited size of that dataset, which consistently resulted in overfitting of the non-CAM proposed model. It is also possible to use the Fourier transform for frequency domain pre-processing \citep{ModelParsing2021}, though it produces complex outputs and is also less efficient.

Experimentation on each dataset is partitioned into \textbf{four phases} of model training followed by evaluation:
\begin{enumerate} [label={\bfseries\Roman*:}]
    \item Deepfake detection, initialization and fitting of all models to clean data.
    \item Deepfake detection, refitting all models to multi-augmented data. Additional instances of each model from phase I are retrained on individually augmented data.
    \item Source attribution, initialization and fitting of baseline models to clean data. Pre-trained instances of the proposed model from phase II are extended with secondary modules, each trained in parallel.
    \item Source attribution, refitting all models to multi-augmented data. New instances of the proposed model are initialized and fitted to individually augmented data using matching primary modules from phase II, whereas the baseline model instances from phase III are retrained accordingly.
\end{enumerate}
In addition to the decentralized binary classification approach used for source attribution, we also conduct conventional multiclass attribution tests exclusively on the GANFP dataset, whereby the baseline models run natively. Our proposed model performs binary classifications in parallel, providing multi-labelled outputs that enable \textit{failed attributions}\footnote{Primary output positive, but all secondary outputs negative.} and \textit{multiple attributions}\footnote{More than one positive secondary output on a given image.}, yielding more information than is previously possible. Multiple independent secondary modules, each branching from the same pre-trained primary module, are initialized and fitted simultaneously to different source GMs. To address compatibility issues when evaluated against the baselines, all failed attributions are interpreted as \texttt{real} predictions in the multiclass attribution scenario. The multiple attribution feature is also suppressed by considering on each image only the secondary output with the highest score.

To measure model performance, we use \textbf{accuracy} for multiclass classification, and \textbf{precision} and \textbf{recall} (including F-scores) for binary classification. The precision value indicates the percentage of correct positive predictions made by the model, whereas the recall (sensitivity) value indicates the percentage of actually relevant images recognized by the model. Relevant images leading to positive predictions are defined as all \texttt{fake} images during deepfake detection, and images from specified sources of interest during attribution. Any sigmoid outputs above the threshold score of 50\% are interpreted as positive predictions. In deepfake detection, although high precision is preferable to avoid unnecessary legal complications, high recall is of utmost importance to minimize the number of synthetic images that would inevitably evade detection. Reduced precision is expected on the augmented datasets since the effects of various degrees and/or combinations of post-processing perturbations may adulterate or eliminate GM fingerprints. For FacesHQ+ experiments, an additional \textbf{external accuracy} metric gauging either the sensitivity (deepfake detection) or specificity (attribution) of the proposed model when encountering novel source GMs intentionally excluded from the training distribution.

\begin{figure*}[h]
    \centering
    \includegraphics[width=\linewidth]{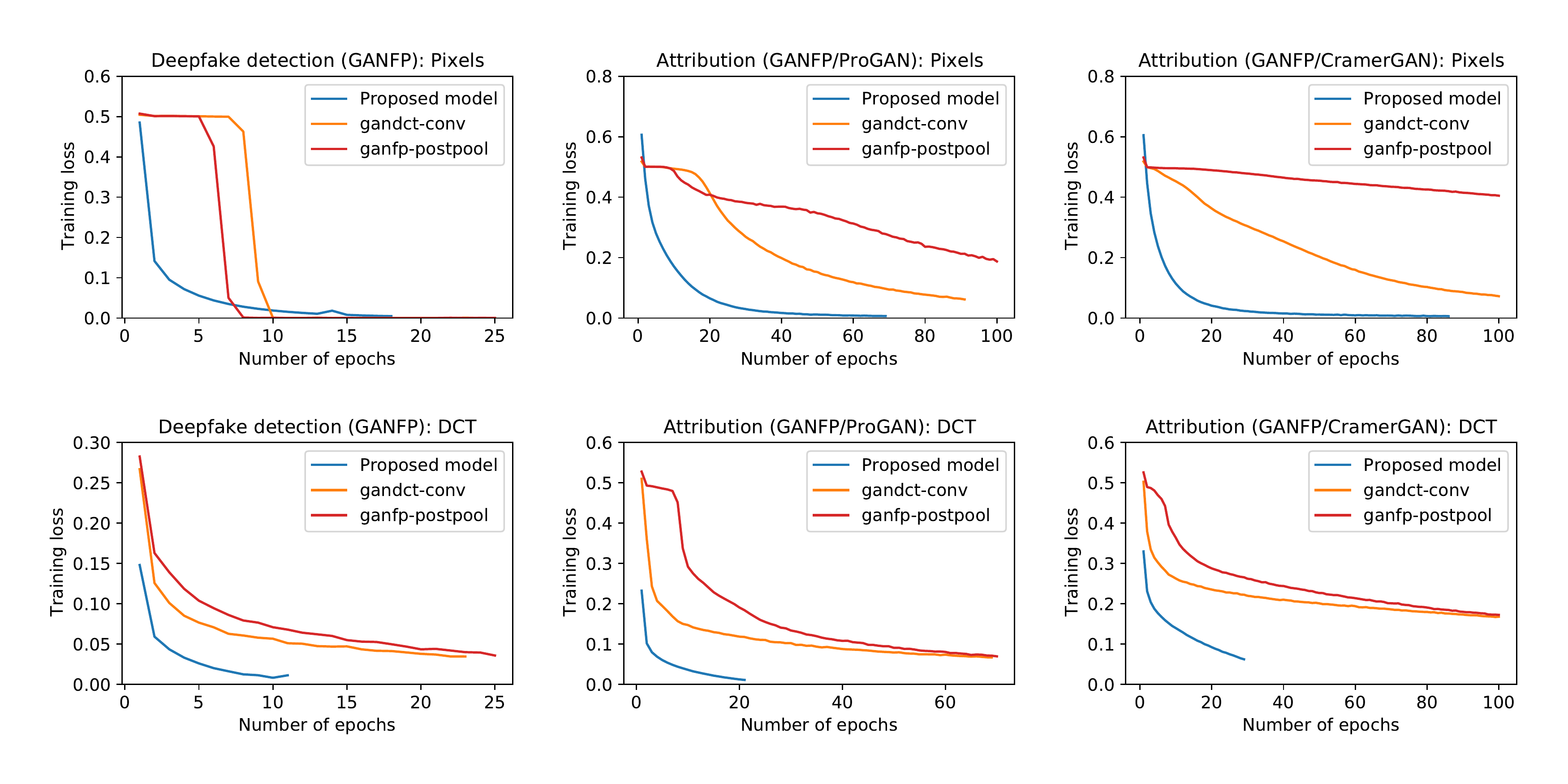}
    \caption[An array of training losses of different classifier models on different GANFP dataset tasks, plotted over the number of training epochs. The proposed model trains considerably faster than the baselines, especially for image source attribution due to transfer learning.]{Training losses over number of epochs, plotted for different classifier models on different GANFP recognition tasks with different input representations. The proposed model is observed to be more efficient relative to both baseline models, especially on source attribution tasks thanks to the application of transfer learning. Model fitting is limited to 100 epochs or until the validation set is overfit, which favours reduced complexity.}
    \label{fig:training_losses}
\end{figure*}

As mentioned in section \ref{framework}, our proposed model also provides feature localization in the form of CAM saliency maps, indicating the observed regions of an input image that heavily influence the prediction made by the model. We evaluated our models' localization function on sets of 16 images taken from each individual image source in the testing datasets. Confidence scores for every prediction are included with the saliency maps. In the binary classification case, the feature activation intensity values of image regions correspond to the degree to which said regions are predicted by the model outputs as containing relevant features. These provide an insight into what image regions or spectral coefficients are recognized by the model as learnt features, particularly when dealing with heavily augmented data or high-level features. However, these saliency maps are semantically distinct from GM fingerprints as the underlying signal-level patterns remain unknown, and therefore should not be considered equivalent to either estimated fingerprints \citep{Yu_2019} or averaged spectral signatures \citep{Wang_2020_ForenSynths}. For maximum visual distinction in otherwise subtle saliency maps, we use the Jet colourmap with low and high feature activations in blue and red respectively.

\noindent Our code is available via the following repository: \textcolor{MidnightBlue}{\url{https://github.com/quarxilon/Generator_Attribution}}.

\section{Experimental Results}

\subsection{Deepfake Detection} \label{deepfake_detection}

\begin{table*}[hp]
\caption[Test results of our evaluated models for deepfake detection.]{Test results of our evaluated models for deepfake detection on the GANFP and FacesHQ+ datasets, before and after augmentation-aware retraining.}

\begin{subtable}{\textwidth}
    \centering
    \caption{GANFP dataset}
    \label{tab:ganfp_det}
    \begin{tabularx}{\textwidth}{|c|X|c c c|c c c|c c c|c c c|}
        \hline
        \multicolumn{2}{|c|}{Test results (\%)} & \multicolumn{6}{c|}{Phase I} & \multicolumn{6}{c|}{Phase II} \\
        \hline
        \multirow{2}{*}{\rotatebox[origin=c]{90}{Input}} & \multirow{2}{*}{Classifier model} & \multicolumn{3}{c|}{CLN set} & \multicolumn{3}{c|}{AUG set} & \multicolumn{3}{c|}{CLN set} & \multicolumn{3}{c|}{AUG set} \\
        & & PRC & REC & F1 & PRC & REC & F1 & PRC & REC & F1 & PRC & REC & F1 \\
        \hline \hline
        \multirow{3}{*}{\rotatebox[origin=c]{90}{PIXEL}}
        & \texttt{gandct-conv}          & 100 & 100 & 100       & 82.0 & 99.2 & 89.8    & 100 & 99.8 & 99.9     & 85.3 & 97.9 & 91.2    \\
        & \texttt{ganfp-postpool}       & 100 & 100 & 100       & 82.9 & 92.7 & 87.6    & 100 & 100 & 100       & 87.1 & 98.3 & 92.4    \\
        \cline{2-14}
        & \textbf{Proposed model}       & 100 & 100 & 100       & 81.9 & 99.7 & 89.9    & 98.4 & 99.8 & 99.1    & 85.0 & 98.0 & 91.1    \\
        \hline \hline
        \multirow{3}{*}{\rotatebox[origin=c]{90}{DCT}}
        & \texttt{gandct-conv}          & 98.7 & 98.0 & 98.9    & 80.1 & 93.3 & 86.2    & 80.0 & 88.4 & 84.0    & 80.0 & 90.0 & 84.7    \\
        & \texttt{ganfp-postpool}       & 99.0 & 99.3 & 99.2    & 85.8 & 48.1 & 61.6    & 97.1 & 99.0 & 98.1    & 87.1 & 98.9 & 92.3    \\
        \cline{2-14}
        & \textbf{Proposed model}       & 98.6 & 99.5 & 99.1    & 83.8 & 78.7 & 81.2    & 96.4 & 98.3 & 97.3    & 85.9 & 96.5 & 90.9    \\
        \hline \hline
        \multicolumn{5}{|l}{\textbf{CLN:} Clean, pristine test dataset} & \multicolumn{9}{l|}{\textbf{PRC:} Precision (True positive rate)} \\
        \multicolumn{5}{|l}{\textbf{AUG:} Multi-augmented test dataset} & \multicolumn{9}{l|}{\textbf{REC:} Recall (Sensitivity)} \\
        \hline
    \end{tabularx}
\end{subtable}

\bigskip \begin{subtable}{\textwidth}
    \centering
    \caption{GANFP dataset, individually augmented}
    \label{tab:ganfp_det_aug}
    \begin{tabularx}{\textwidth}{|c|X|c c c|c c c|c c c|c c c|}
        \hline
        \multicolumn{2}{|c|}{Test results (\%)} & \multicolumn{6}{c|}{Phase I} & \multicolumn{6}{c|}{Phase II} \\
        \hline
        \multirow{2}{*}{\rotatebox[origin=c]{90}{Input}} & \multirow{2}{*}{Classifier model} & \multicolumn{3}{c|}{JPEG set} & \multicolumn{3}{c|}{CROP set} & \multicolumn{3}{c|}{JPEG set} & \multicolumn{3}{c|}{CROP set} \\
        & & PRC & REC & F1 & PRC & REC & F1 & PRC & REC & F1 & PRC & REC & F1 \\
        \hline \hline
        \multirow{3}{*}{\rotatebox[origin=c]{90}{PIXEL}}
        & \texttt{gandct-conv}          & 100 & 100 & 100       & 88.9 & 97.4 & 93.0    & 100 & 100 & 100       & 100 & 100 & 100   \\
        & \texttt{ganfp-postpool}       & 100 & 93.7 & 96.7     & 96.8 & 86.8 & 91.5    & 100 & 100 & 100       & 100 & 100 & 100   \\
        \cline{2-14}
        & \textbf{Proposed model}       & 98.4 & 100 & 99.2     & 88.7 & 98.9 & 93.5    & 100 & 100 & 100       & 100 & 100 & 100   \\
        \hline \hline
        \multirow{3}{*}{\rotatebox[origin=c]{90}{DCT}}
        & \texttt{gandct-conv}          & 98.5 & 98.6 & 98.6    & 98.2 & 65.9 & 78.9    & 98.7 & 98.8 & 98.7    & 96.8 & 98.8 & 97.8    \\
        & \texttt{ganfp-postpool}       & 98.8 & 99.2 & 99.0    & 98.8 & 58.4 & 73.4    & 99.2 & 99.3 & 99.2    & 98.2 & 98.9 & 98.5    \\
        \cline{2-14}                                                                                                          
        & \textbf{Proposed model}       & 98.4 & 99.4 & 98.9    & 92.0 & 88.6 & 90.3    & 98.9 & 99.4 & 99.2    & 98.2 & 98.7 & 98.4    \\
        \hline \hline                                                                                                        
        \multicolumn{5}{|l}{\textbf{JPEG:} JPEG compressed test dataset} & \multicolumn{9}{l|}{\textbf{CROP:} Randomly cropped test dataset} \\
        \hline
    \end{tabularx}
\end{subtable}

\bigskip \begin{subtable}{\textwidth}
    \centering \setlength{\tabcolsep}{10pt}
    \caption{FacesHQ+ dataset}
    \label{tab:faceshqp_det}
    \begin{tabularx}{\textwidth}{|X|c c c c|c c c c|}
        \hline
        \textbf{Proposed model (Pixel)} & \multicolumn{4}{c|}{Phase I} & \multicolumn{4}{c|}{Phase II} \\
        \hline
        Test results (\%) & PRC & REC & F1 & EXA & PRC & REC & F1 & EXA \\
        \hline \hline
        Clean test set                  & 90.7 & 91.4 & 91.0 & 10.6     & 90.4 & 88.3 & 89.3 & 14.3    \\
        Multi-augmented test set        & 88.1 & 43.4 & 58.2 & 9.8      & 84.5 & 85.8 & 85.2 & 29.6    \\
        \hline \hline
        JPEG compressed test set        & 90.7 & 91.4 & 91.0 & 10.6     & 88.9 & 91.8 & 90.3 & 13.9    \\
        Randomly cropped test set       & 90.2 & 70.2 & 79.0 & 12.0     & 88.9 & 89.1 & 89.0 & 22.2    \\
        \hline \hline
        \multicolumn{9}{|l|}{\textbf{EXA:} External accuracy, i.e. Recall on StarGANv2 test set} \\
        \hline
    \end{tabularx}
\end{subtable}
\end{table*}

Firstly, all three classifier models including their DCT spectral input variants are trained on the GANFP dataset (100,000 images/epoch) for deepfake detection. Table \ref{tab:ganfp_det} lists the test results of the models before and after retraining on multi-augmented data\footnote{Data randomly subjected to all four augmentations described in section \ref{datasets}.} (phases I and II respectively). In this case, the proposed model only consists of its primary module and single output, which fitted the dataset to satisfactory levels slightly faster than baseline as stated in \ref{fig:training_losses}. Source attribution tests later in phases III and IV considerably enhanced the rapid training advantage of the proposed model once its primary module is sufficiently fitted, since the secondary modules have access to appropriately processed features from the start, despite the overall model being ultimately limited by the errors of the primary module. Regardless, the developmental efficiency benefits of adopting transfer learning in deepfake forensics are conspicuous, at least on lower-resolution and less varied datasets.

We tested the models on both clean and multi-augmented data, expecting reduced accuracy on the latter as with other studies. While all models performed almost equally well, the DCT spectral input variants found the multi-augmented data challenging. The results also imply that the \texttt{ganfp-postpool} model is more capable of adapting to additive perturbations than its less complex counterparts \citep{Frank_2020}. Additionally, the lower precision values recorded on the multi-augmented set corroborate the overlap between the learnt synthetic image fingerprints and intense post-processing artefacts. Similarly, table \ref{tab:ganfp_det_aug} lists the test results of the models before and after retraining on individually augmented datasets \citep{mandelli2020training}, namely JPEG compression and random asymmetric cropping. Retraining on compressed images is found to have negligible effects, whereas the DCT input variants are somewhat vulnerable to image cropping. Performance on both individually augmented sets exceed those of the multi-augmented set, implying that the blurring (low-pass filter) and additive noise affect the models more severely.

The performance of the proposed model on the high-resolution FacesHQ+ dataset is reported in table \ref{tab:faceshqp_det}, including the \textit{external accuracy} (EXA) metric for out-of-distribution StarGANv2 samples. The CAM structure of the proposed model allows it to accept 256x256 image inputs without any modifications, while early stopping regularization is relaxed to 10 epochs. Nonetheless, it is more difficult for the model to tell StyleGAN/StyleGAN2 images apart from `real' images. Several factors are at play here: the small size of the FacesHQ+ dataset limits the range of signal-level features that can comprise reliable fingerprints, which are themselves barely perceptible in the outputs of the technically advanced, highly-developed GMs of the StyleGAN family; and thus more likely to be extinguished during image rescaling and post-processing. As with GANFP, phase II retraining on individually augmented data boosts recall on said data but provides no other major improvements.

What is more concerning is that deepfake detection EXA (i.e. recall) is far worse than the expected 50\% chance when randomly classifying an entirely fake image set. Admittedly, the StarGANv2 images that comprise the out-of-distribution set are created using a conceptually different style transfer GAN that conditionally edits real images while retaining certain visual information. However, the results show that these images consistently fail to be flagged by the proposed model in its current design, suggesting that the fingerprints learnt by the model are limited to the scope of the FacesHQ+ training dataset and therefore not externally valid. This finding may potentially be invalidated with more diverse training datasets and repurposing of deeper classifier models with better generalization ability as primary modules. Previously, \citet{Wang_2020_ForenSynths} optimized an ImageNet-pretrained ResNet-50 on a large multi-domain assortment of only ProGAN-generated images, which was subsequently able to generalize effectively to various types of GMs including auto-encoded deepfake videos. Interestingly, phase II retraining with augmentations marginally increases external accuracy.

\subsection{Image Source Attribution} \label{source_attribution}

\begin{table*}[hp]
\caption[Test results of our evaluated models for binary image source attribution]{Test results of our evaluated models for binary image source attribution on the GANFP and FacesHQ+ datasets, before and after augmentation-aware retraining.}

\begin{subtable}{\textwidth}
    \centering
    \caption{GANFP dataset, \textbf{ProGAN} as source GM of interest.}
    \label{tab:ganfp_att_progan}
    \begin{tabularx}{\textwidth}{|c|X|c c c|c c c|c c c|c c c|}
        \hline
        \multicolumn{2}{|c|}{Test results (\%)} & \multicolumn{6}{c|}{Phase III} & \multicolumn{6}{c|}{Phase IV} \\
        \hline
        \multirow{2}{*}{\rotatebox[origin=c]{90}{Input}} & \multirow{2}{*}{Classifier model} & \multicolumn{3}{c|}{CLN set} & \multicolumn{3}{c|}{AUG set} & \multicolumn{3}{c|}{CLN set} & \multicolumn{3}{c|}{AUG set} \\
        & & PRC & REC & F1 & PRC & REC & F1 & PRC & REC & F1 & PRC & REC & F1 \\
        \hline \hline
        \multirow{3}{*}{\rotatebox[origin=c]{90}{PIXEL}}
        & \texttt{gandct-conv}      & 93.7 & 91.0 & 92.4    & \tbf{67.6} & \tbf{36.0} & \tbf{47.0}  & 88.2 & 84.0 & 86.0    & 73.9 & \tbf{63.7} & \tbf{68.4}    \\
        & \texttt{ganfp-postpool}   & 84.3 & 80.8 & 82.5    & 47.1 & 32.3 & 38.3                    & 77.3 & 62.3 & 69.0    & 76.4 & 37.1 & 50.0                \\
        \cline{2-14}
        & \textbf{Proposed model}   & \tbf{99.2} & \tbf{99.5} & \tbf{99.4}  & 40.5 & 24.9 & 30.8    & \tbf{99.1} & \tbf{87.6} & \tbf{93.0}  & \tbf{83.8} & 15.2 & 25.7  \\
        \hline \hline
        \multirow{3}{*}{\rotatebox[origin=c]{90}{DCT}}
        & \texttt{gandct-conv}      & 93.6 & 90.9 & 92.2    & 42.0 & \tbf{23.7} & \tbf{30.3}        & 89.6 & 77.6 & 83.2    & 75.8 & 25.8 & 38.5                \\
        & \texttt{ganfp-postpool}   & 91.5 & 92.0 & 91.8    & \tbf{67.5} & 16.1 & 26.0              & 91.2 & 79.1 & 84.7    & \tbf{81.6} & 23.3 & 36.2          \\
        \cline{2-14}
        & \textbf{Proposed model}   & \tbf{94.7} & \tbf{93.8} & \tbf{94.3}  & 65.7 & 17.5 & 27.7    & \tbf{92.0} & \tbf{80.4} & \tbf{85.8}  & 73.4 & \tbf{26.2} & \tbf{38.6}    \\
        \hline \hline
        \multicolumn{5}{|l}{\textbf{CLN:} Clean, pristine test dataset} & \multicolumn{9}{l|}{\textbf{PRC:} Precision (True positive rate)} \\
        \multicolumn{5}{|l}{\textbf{AUG:} Multi-augmented test dataset} & \multicolumn{9}{l|}{\textbf{REC:} Recall (Sensitivity)} \\
        \hline
    \end{tabularx}
\end{subtable}

\bigskip \begin{subtable}{\textwidth}
    \centering
    \caption{GANFP dataset, \textbf{CramerGAN} as source GM of interest.}
    \label{tab:ganfp_att_cramergan}
    \begin{tabularx}{\textwidth}{|c|X|c c c|c c c|c c c|c c c|}
        \hline
        \multicolumn{2}{|c|}{Test results (\%)} & \multicolumn{6}{c|}{Phase III} & \multicolumn{6}{c|}{Phase IV} \\
        \hline
        \multirow{2}{*}{\rotatebox[origin=c]{90}{Input}} & \multirow{2}{*}{Classifier model} & \multicolumn{3}{c|}{CLN set} & \multicolumn{3}{c|}{AUG set} & \multicolumn{3}{c|}{CLN set} & \multicolumn{3}{c|}{AUG set} \\
        & & PRC & REC & F1 & PRC & REC & F1 & PRC & REC & F1 & PRC & REC & F1 \\
        \hline \hline
        \multirow{3}{*}{\rotatebox[origin=c]{90}{PIXEL}}
        & \texttt{gandct-conv}      & 90.0 & 89.2 & 89.6    & \tbf{65.1} & \tbf{46.6} & \tbf{54.4}      & 87.6 & 87.3 & 87.4    & \tbf{77.1} & \tbf{64.0} & \tbf{70.0}      \\
        & \texttt{ganfp-postpool}   & 63.2 & 23.1 & 33.8    & 58.8 & 12.2 & 20.2                        & 61.5 & 39.5 & 48.1    & 59.6 & 28.4 & 38.5                        \\
        \cline{2-14}
        & \textbf{Proposed model}   & \tbf{98.9} & \tbf{98.7} & \tbf{98.8}      & 44.0 & 18.6 & 26.2    & \tbf{92.5} & \tbf{95.5} & \tbf{94.0}      & 73.3 & 18.3 & 29.3    \\
        \hline \hline
        \multirow{3}{*}{\rotatebox[origin=c]{90}{DCT}}
        & \texttt{gandct-conv}      & \tbf{81.9} & 74.0 & 77.8          & 56.2 & 15.0 & 23.7                & 72.4 & 57.3 & 64.0                    & 66.4 & 14.2 & 23.4                    \\
        & \texttt{ganfp-postpool}   & 77.3 & \tbf{82.5} & \tbf{79.8}    & 34.0 & \tbf{21.1} & \tbf{26.1}    & \tbf{77.2} & \tbf{73.5} & \tbf{75.3}  & \tbf{71.9} & \tbf{21.2} & \tbf{32.7}  \\
        \cline{2-14}
        & \textbf{Proposed model}   & 80.4 & 76.8 & 78.6    & \tbf{56.4} & 15.0 & 23.7      & 75.1 & 65.3 & 69.9    & 62.8 & 16.1 & 25.6    \\
        \hline
    \end{tabularx}
\end{subtable}

\bigskip \begin{subtable}{\textwidth}
    \centering \setlength{\tabcolsep}{10pt}
    \caption{FacesHQ+ dataset}
    \label{tab:faceshqp_att}
    \begin{tabularx}{\textwidth}{|X|c c c c|c c c c|}
        \hline
        \textbf{Proposed model (Pixel)} & \multicolumn{4}{c|}{Phase III} & \multicolumn{4}{c|}{Phase IV} \\
        \hline
        Test results (\%) & PRC & REC & F1 & EXA & PRC & REC & F1 & EXA \\
        \hline \hline
        \multicolumn{9}{|c|}{\textbf{StyleGAN (ThisPersonDoesNotExist)} Instance level attribution}     \\
        \hline
        Clean test set                  & 75.9 & 33.5 & 46.5 & 99.8     & 74.8 & 35.2 & 47.8 & 99.8     \\
        Multi-augmented test set        & 49.3 & 22.9 & 31.2 & 98.0     & 66.5 & 23.1 & 34.3 & 99.1     \\
        \hline \hline
        \multicolumn{9}{|c|}{\textbf{StyleGAN2} Model level attribution} \\
        \hline
        Clean test set                  & 83.0 & 50.3 & 62.6 & 99.3     & 75.3 & 56.3 & 64.4 & 98.2     \\
        Multi-augmented test set        & 80.3 & 16.1 & 26.8 & 99.4     & 70.6 & 40.8 & 51.7 & 97.0     \\
        \hline \hline
        \multicolumn{9}{|l|}{\textbf{EXA:} External accuracy, i.e. Specificity on StarGANv2 test set}   \\
        \hline
    \end{tabularx}
\end{subtable}
\end{table*}

\begin{table*}[hp]
\caption[Test results of our evaluated models for binary image source attribution (individual augmentations)]{Test results of our evaluated models for binary image source attribution on individually augmented versions of the GANFP and FacesHQ+ datasets. The secondary layers of the proposed models are retrained from scratch with specialized primary modules.}

\begin{subtable}{\textwidth}
    \centering
    \caption{GANFP dataset, \textbf{ProGAN} as source GM of interest.}
    \label{tab:ganfp_att_progan_aug}
    \begin{tabularx}{\textwidth}{|c|X|c c c|c c c|c c c|c c c|}
        \hline
        \multicolumn{2}{|c|}{Test results (\%)} & \multicolumn{6}{c|}{Phase III} & \multicolumn{6}{c|}{Phase IV} \\
        \hline
        \multirow{2}{*}{\rotatebox[origin=c]{90}{Input}} & \multirow{2}{*}{Classifier model} & \multicolumn{3}{c|}{JPEG set} & \multicolumn{3}{c|}{CROP set} & \multicolumn{3}{c|}{JPEG set} & \multicolumn{3}{c|}{CROP set} \\
        & & PRC & REC & F1 & PRC & REC & F1 & PRC & REC & F1 & PRC & REC & F1 \\
        \hline \hline
        \multirow{3}{*}{\rotatebox[origin=c]{90}{PIXEL}}
        & \texttt{gandct-conv}          & 93.1 & 86.9 & 89.9    & \tbf{86.0} & 63.8 & 73.2          & 92.6 & \tbf{91.8} & 92.2      & 87.5 & 85.3 & 86.4    \\
        & \texttt{ganfp-postpool}       & 85.6 & 51.6 & 64.4    & 83.6 & \tbf{79.2} & \tbf{81.3}    & 85.0 & 71.6 & 77.7            & 93.0 & 88.6 & 90.8    \\
        \cline{2-14}
        & \textbf{Proposed model}       & \tbf{96.5} & \tbf{91.3} & \tbf{93.8}      & 60.4 & 78.6 & 68.3        & \tbf{94.6} & 91.3 & \tbf{93.0}        & \tbf{95.8} & \tbf{88.7} & \tbf{92.1}  \\
        \hline \hline
        \multirow{3}{*}{\rotatebox[origin=c]{90}{DCT}}
        & \texttt{gandct-conv}          & 92.3 & 87.5 & 89.9    & 74.1 & 51.2 & 60.6                & 92.9 & 88.9 & 90.8                & 86.0 & 79.6 & 82.7    \\
        & \texttt{ganfp-postpool}       & 88.8 & 89.3 & 89.1    & \tbf{77.5} & 49.9 & 60.7          & 93.0 & \tbf{90.2} & \tbf{91.6}    & 88.0 & 85.6 & 86.8    \\
        \cline{2-14}
        & \textbf{Proposed model}       & \tbf{92.9} & \tbf{90.0} & \tbf{91.4}      & 66.0 & \tbf{56.5} & \tbf{60.9}    & \tbf{93.6} & 89.5 & 91.5      & \tbf{93.6} & \tbf{89.5} & \tbf{91.5}  \\
        \hline \hline
        \multicolumn{5}{|l}{\textbf{JPEG:} JPEG compressed test dataset}    & \multicolumn{9}{l|}{\textbf{PRC:} Precision (True positive rate)} \\
        \multicolumn{5}{|l}{\textbf{CROP:} Randomly cropped test dataset}   & \multicolumn{9}{l|}{\textbf{REC:} Recall (Sensitivity)} \\
        \hline
    \end{tabularx}
\end{subtable}

\bigskip \begin{subtable}{\textwidth}
    \centering
    \caption{GANFP dataset, \textbf{CramerGAN} as source GM of interest.}
    \label{tab:ganfp_att_cramergan_aug}
    \begin{tabularx}{\textwidth}{|c|X|c c c|c c c|c c c|c c c|}
        \hline
        \multicolumn{2}{|c|}{Test results (\%)} & \multicolumn{6}{c|}{Phase III} & \multicolumn{6}{c|}{Phase IV} \\
        \hline
        \multirow{2}{*}{\rotatebox[origin=c]{90}{Input}} & \multirow{2}{*}{Classifier model} & \multicolumn{3}{c|}{JPEG set} & \multicolumn{3}{c|}{CROP set} & \multicolumn{3}{c|}{JPEG set} & \multicolumn{3}{c|}{CROP set} \\
        & & PRC & REC & F1 & PRC & REC & F1 & PRC & REC & F1 & PRC & REC & F1 \\
        \hline \hline
        \multirow{3}{*}{\rotatebox[origin=c]{90}{PIXEL}}
        & \texttt{gandct-conv}      & 90.5 & \tbf{86.8} & \tbf{88.6}    & 69.0 & \tbf{66.0} & 67.5          & 90.7 & \tbf{90.3} & \tbf{90.5}    & 83.0 & 76.2 & 79.5    \\
        & \texttt{ganfp-postpool}   & 63.0 & 22.2 & 32.8                & 59.3 & 21.2 & 31.2                & 71.7 & 37.6 & 49.4                & 65.2 & 35.2 & 45.7    \\
        \cline{2-14}
        & \textbf{Proposed model}   & \tbf{98.6} & 64.9 & 78.3          & \tbf{97.6} & 49.7 & \tbf{65.9}    & \tbf{98.1} & 83.3 & 90.1          & \tbf{99.5} & \tbf{97.1} & \tbf{98.3}  \\
        \hline \hline
        \multirow{3}{*}{\rotatebox[origin=c]{90}{DCT}}
        & \texttt{gandct-conv}      & \tbf{82.5} & 65.2 & 72.8          & \tbf{80.2} & 37.9 & 51.5          & 79.3 & 73.1 & 76.1                & 69.9 & 58.7 & 63.8                \\
        & \texttt{ganfp-postpool}   & 77.1 & \tbf{76.5} & \tbf{76.8}    & 77.2 & \tbf{41.5} & \tbf{54.0}    & 81.8 & \tbf{79.7} & \tbf{80.7}    & \tbf{83.5} & 61.4 & \tbf{70.8}    \\
        \cline{2-14}
        & \textbf{Proposed model}   & 80.6 & 69.3 & 74.6                & 78.0 & 39.4 & 52.3                & \tbf{81.9} & 69.8 & 75.4          & 73.2 & \tbf{62.4} & 67.4          \\
        \hline
    \end{tabularx}
\end{subtable}

\bigskip \begin{subtable}{\textwidth}
    \centering \setlength{\tabcolsep}{10pt}
    \caption{FacesHQ+ dataset}
    \label{tab:faceshqp_att_aug}
    \begin{tabularx}{\textwidth}{|X|c c c c|c c c c|}
        \hline
        \textbf{Proposed model (Pixel)} & \multicolumn{4}{c|}{Phase III} & \multicolumn{4}{c|}{Phase IV} \\
        \hline
        Test results (\%) & PRC & REC & F1 & EXA & PRC & REC & F1 & EXA \\
        \hline \hline
        \multicolumn{9}{|c|}{\textbf{StyleGAN (ThisPersonDoesNotExist)} Instance level attribution} \\
        \hline
        JPEG compressed test set         & 76.5 & 33.5 & 46.6 & 99.8    & 78.7 & 27.7 & 41.0 & 99.9     \\
        Randomly cropped test set        & 73.2 & 22.1 & 34.0 & 99.4    & 74.1 & 31.9 & 44.6 & 98.4     \\
        \hline \hline
        \multicolumn{9}{|c|}{\textbf{StyleGAN2} Model level attribution} \\
        \hline
        JPEG compressed test set         & 82.8 & 50.5 & 62.7 & 99.4    & 93.9 & 21.7 & 35.2 & 100      \\
        Randomly cropped test set        & 81.6 & 36.0 & 50.0 & 99.1    & 83.3 & 36.2 & 50.5 & 99.4     \\
        \hline \hline
        \multicolumn{9}{|l|}{\textbf{EXA:} External accuracy, i.e. Specificity on StarGANv2 test set} \\
        \hline
    \end{tabularx}
\end{subtable}
\end{table*}

For binary-class attribution, we selected the following source GMs as attribution targets: \textbf{ProGAN} \& \textbf{CramerGAN} for GANFP, and \textbf{StyleGAN} \& \textbf{StyleGAN2} for FacesHQ+. Both our baseline classifiers (Figure \ref{fig:baselines_diagram}) can be directly fitted for binary-class attribution with a sigmoid output layer, whereas our proposed models make use of trainable secondary modules branching off primary modules adapted from table \ref{tab:ganfp_det}, pre-acclimatized on multi-augmented data. Table \ref{tab:ganfp_att_progan} lists the test results of the model instances for ProGAN attribution. The proposed models achieved results equivalent to (or marginally better than) baselines when evaluated on the clean test set; however, the baselines fared better against multi-augmented data. Note that recall scores plummet across the board when faced with multi-augmented data; this effect also extends to clean test data after phase IV. The clean self-contradiction rate (CR) of the proposed model is around 0.5\% (pixel) or 2.3\% (DCT); multi-augmented data caused the CR to either increase (pixel) or decrease (DCT).

\begin{table*}[h]
\caption[Test results of our evaluated models on the GANFP dataset for multiclass image source attribution. The proposed models actually perform multiple binary classifications in parallel.]{Test results of our models on the GANFP dataset for multiclass image source attribution, before and after augmentation-aware retraining. The proposed models perform multiple binary classifications in parallel.}
\centering \setlength{\tabcolsep}{10pt}
\label{tab:ganfp_att_multiclass}
\begin{tabularx}{\textwidth}{|c|X|c c|c c|c c|c c|}
    \hline
    \multirow{2}{*}{\rotatebox[origin=c]{90}{Input}} & \textbf{Accuracy} (\%) & \multicolumn{4}{c|}{Phase III} & \multicolumn{4}{c|}{Phase IV} \\
    \cline{2-10}
    & Classifier model & CLN & AUG & JPEG & CROP & CLN & AUG & JPEG & CROP \\
    \hline \hline
    \multirow{3}{*}{\rotatebox[origin=c]{90}{PIXEL}}
    & \texttt{gandct-conv}          & 93.5 & \tbf{47.7} & \tbf{89.9} & 63.1         & \tbf{89.3} & 60.9 & \tbf{91.5} & 83.8 \\
    & \texttt{ganfp-postpool}       & 91.1 & 45.7 & 86.6 & \tbf{81.9}               & 87.9 & \tbf{73.1} & 88.3 & 90.5 \\
    \cline{2-10}
    & \textbf{Proposed model}       & \tbf{96.3} & 28.5 & 79.5 & 57.0               & 84.0 & 29.5 & 85.0 & \tbf{95.3} \\
    \hline \hline
    \multirow{3}{*}{\rotatebox[origin=c]{90}{DCT}}
    & \texttt{gandct-conv}          & \tbf{91.4} & \tbf{32.6} & \tbf{89.2} & \tbf{58.9}     & 81.3 & 45.4 & \tbf{90.1} & 83.3 \\
    & \texttt{ganfp-postpool}       & 88.6 & 30.5 & 86.3 & 55.8                             & \tbf{83.5} & \tbf{47.8} & 88.3 & \tbf{84.4} \\
    \cline{2-10}
    & \textbf{Proposed model}       & 85.9 & 30.9 & 82.4 & 54.8                             & 72.4 & 36.9 & 83.7 & 77.1 \\
    \hline \hline
    \multicolumn{4}{|l}{\textbf{CLN:} Clean, pristine test dataset} & \multicolumn{6}{l|}{\textbf{JPEG:} JPEG compressed test dataset} \\
    \multicolumn{4}{|l}{\textbf{AUG:} Multi-augmented test dataset} & \multicolumn{6}{l|}{\textbf{CROP:} Randomly cropped test dataset} \\
    \hline
\end{tabularx}
\end{table*}

Similarly, table \ref{tab:ganfp_att_cramergan} lists the test results of the CramerGAN attribution models. The additional challenge provided by this particular implementation of the eponymous GM is that it shares most of its codebase with the MMDGAN implementation \citep{Yu_2019}, and therefore manifests somewhat similar fingerprints influenced by shared GM hyperparameters. This challenge makes approaching image source attribution as a binary classification problem more difficult than in multiclass classification. Our observations indicate a considerable drop in discriminative performance (especially recall) across all models, implying that the models are forced to identify low-attributability fingerprints not shared between CramerGAN and MMDGAN. In fact, expecting increased training difficulty, early stopping regularization is relaxed to 20 epochs; our baselines still hit the 100 epoch limit before converging. The clean CR of the proposed model this time is around 0.1\% (pixel) or 2.8\% (DCT).

Tables \ref{tab:ganfp_att_progan_aug} \& \ref{tab:ganfp_att_cramergan_aug} list the individual augmentation test results of the ProGAN and CramerGAN attribution models from tables \ref{tab:ganfp_att_progan} \& \ref{tab:ganfp_att_cramergan} respectively. Our models' primary modules are replaced with those evaluated in table \ref{tab:ganfp_det_aug} affixed to new secondary layers fitted from scratch, whereas the baseline models are retrained as normal. Apparently, the forensic challenges brought on by JPEG compression and random cropping on their own are far less pronounced. We suspect that some of the heavier augmentations present in the multi-augmented set (e.g. blurred images) thoroughly eliminate intrinsic fingerprints usable for attribution, thus severely limiting the attainable true positive rate. Incidentally, all attribution models trained from scratch experienced 1--2 epochs of stagnancy shortly after initialization, where the cross-entropy loss would continue to decrease while training accuracy remains exactly at the ratio of negative labels to positive labels. This corresponds to the situation where the models simply predict \emph{all} images as negative samples from $\cl{G}_{\phi}^{\prime}$ (which would still result in high accuracy) instead of finding relevant, distinguishable fingerprints from $G_{\phi}$.

We conducted multiclass attribution tests on the GANFP dataset, though the test accuracies reported in table \ref{tab:ganfp_att_multiclass} are slightly worse than those by \citet{Frank_2020}. We suspect three plausible reasons: reduced dataset size (20\% of the original), differing model hyperparameters (learning rate and na\"{i}ve initialization), and limited model fitting. Although the baselines appear to outperform the proposed model, the latter is actually performing multiple binary classifications in parallel; hence, it lacks the support of mutual exclusivity. Despite the increased classification difficulty dealt to the proposed model, it is nearly as competent as both baselines. All models (especially their DCT variants) have difficulty differentiating between CramerGAN and MMDGAN images, whereas multi-augmented fake images are frequently predicted as \texttt{real} (implying significant fingerprint erasure). The CR of our model ranges from 1--8\% on clean data, but reach 25--45\% on multi-augmented data. Augmentation-aware retraining reduces contradictory outputs for pixel inputs, but not DCT inputs.

For FacesHQ+, we experimented with source attribution at the architectural and instance levels. Both selected source GMs (StyleGAN and StyleGAN2) were hosted on \texttt{ThisPersonDoesNotExist.com}. Instance-level attribution leads to conflict with StyleGAN images from another source (\texttt{generated.photos}), which is evident in tables \ref{tab:faceshqp_att} \& \ref{tab:faceshqp_att_aug} where StyleGAN (TPDNE) attribution leads to worse outcomes than StyleGAN2 attribution. Unfortunately, recall values are considerably lower in FacesHQ+ tests compared to GANFP; with the majority of positively labelled images fail to be recognized regardless of applied augmentations. This could be due to either insufficient training data, insufficient classifier capacity, or erasure of critical features during image downsampling. Moreover, the primary modules fitted to FacesHQ+ are relatively suboptimal, compromising the quality of extracted features used during secondary module training. The StyleGAN2-focused model instance in table \ref{tab:faceshqp_att_aug} achieved a CR under 1\%, which otherwise ranges between 2--4\%. Despite mediocre model sensitivity, the secondary modules consistently reject StarGANv2-sourced out-of-distribution images, as exemplified by the high source attribution EXA values (i.e. specificity). This presents an interesting observation: the model is unable to recognize StarGANv2 images as \texttt{fake}, but manages to satisfy its secondary objective with minimal misattribution of images from an unknown source.

\subsection{Localization and Interpretation}

\begin{figure*}[ht]
    \centering
    \includegraphics[width=0.9\linewidth]{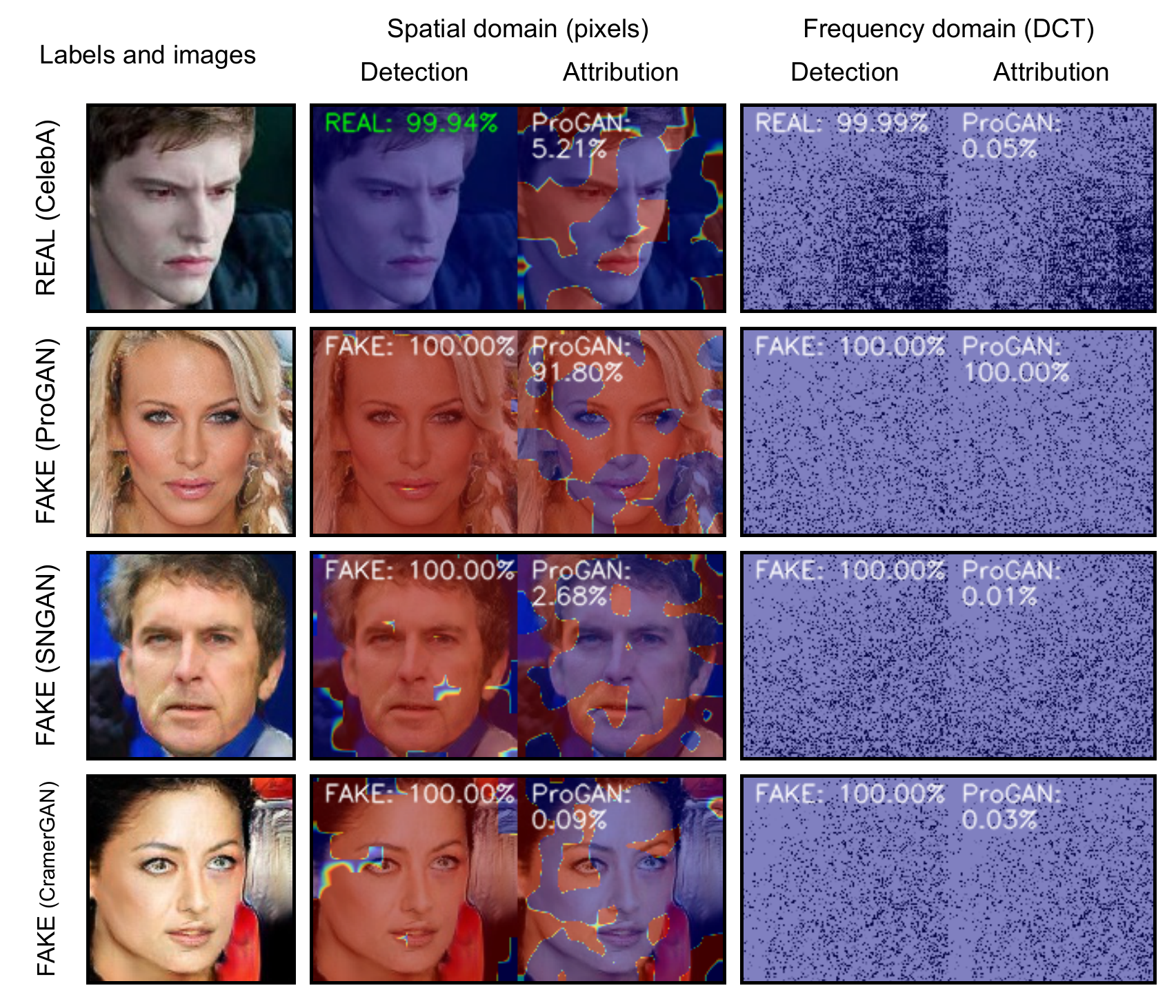}
    \caption[Class activation saliency maps by our model for feature localization of deepfake detection and source attribution of ProGAN images in the GANFP dataset.]{Class activation saliency maps (pixel and DCT analysis respectively) by our model trained to phase IV on the individually augmented GANFP dataset (JPEG compression) for feature localization of deepfake detection and source attribution of ProGAN images, based on results from table \ref{tab:ganfp_att_progan_aug}. All images depicted here are compressed and have their respective sources stated to the left. Model predictions and confidence scores are superimposed on each saliency map.}
    \label{fig:localization_samples}
\end{figure*}

When instances of our feature localization-enabled model are initially fitted to GANFP images, inactivated regions are observed throughout \texttt{real} samples though spikes do occur to a mostly negligible degree, whereas high activations are prevalent within the foregrounds of \texttt{fake} samples. As of training phase III, the highly activated regions of secondary outputs for source attribution are almost always distributed without obvious consistent patterns, while their coverage and distribution correlate less with the associated confidence scores compared to the primary outputs. On DCT frequency spectra, the coefficients leading to class activations are concentrated at lower bands (upper left) for \texttt{real} images but distributed evenly across the spectrum for \texttt{fake} images; thus corroborating prior findings concerning the significance of higher-frequency components to deepfake detection. Despite that, occasional spikes in \texttt{real} images are enough to result in misclassification. It is still unclear if these are due to premature model fitting or existing perturbations in the supposedly clean images. We also found increased model confusion between CramerGAN and MMDGAN sources in the frequency domain.

\begin{figure}[h]
    \centering
    \includegraphics[width=\linewidth]{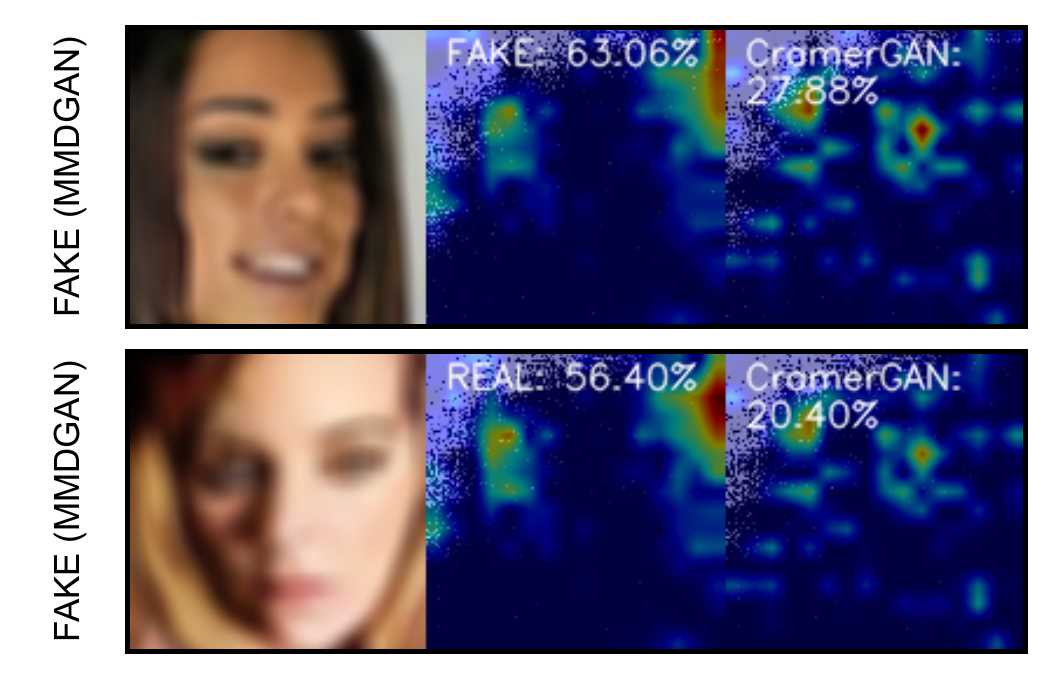}
    \caption[DCT spectral Grad-CAM maps of two similarly augmented MMDGAN images, one of which is misclassified by our model.]{DCT spectral Grad-CAMs of two MMDGAN-sourced images, from the outputs of our phase IV retrained model for attributing CramerGAN images. Both images originated from the same source and are heavily corrupted with Gaussian blurring and random cropping, but the second image is somehow misclassified as \texttt{real}.}
    \label{fig:localization_weirdness}
\end{figure}

As expected, multi-augmented images present major challenges to our models even after phase II/IV retraining. Classification errors are rampant with almost all augmented \texttt{real} images being predicted as \texttt{fake} with high confidence and extensive activated regions, whereas \texttt{fake} images are mostly recognized (but not attributed) correctly albeit with conspicuously low activations surrounding human eyes. Only clean images are positively attributed with (mostly) high scores, while augmentations cause negative attributions of varying scores \textit{regardless of source}. Gaussian blurring in particular is associated with high image-wide activations across all images for both pixel and DCT model variants. Interestingly, two similarly augmented MMDGAN images (Figure \ref{fig:localization_weirdness}) have similar topographical distributions within their DCT spectral activation maps, but one of them is misclassified as \texttt{real}. In general, blurring and additive noise appear to severely hinder frequency analysis of synthetic images through the elimination and fabrication of high-frequency components respectively, thus resulting in frequent binary classification errors during source attribution via spectral analysis.

\begin{figure*}[ht]
    \centering
    \includegraphics[width=\linewidth]{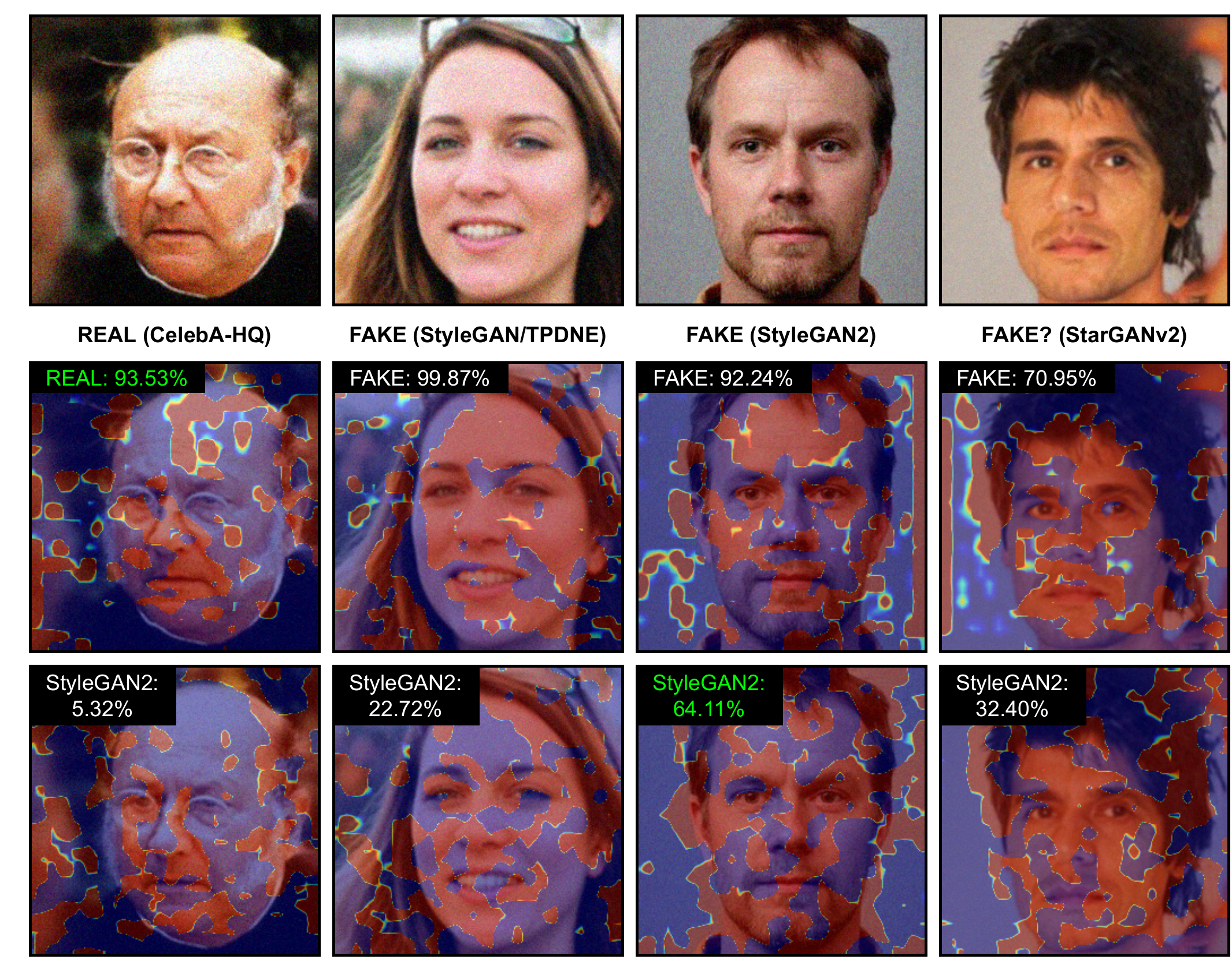}
    \caption[Class activation maps by our model for feature localization of deepfake detection and source attribution of multi-augmented StyleGAN2 images in the FacesHQ+ dataset.]{Class activation maps by our model trained to phase IV on the multi-augmented FacesHQ+ dataset for feature localization of deepfake detection and source attribution of StyleGAN2 images, based on results from table \ref{tab:faceshqp_att}. All images depicted here are heavily degraded with at least three consecutive augmentations. The StarGANv2 image falls outside the training distribution and used to estimate external validity.}
    \label{fig:localization_samples_faceshqp}
\end{figure*}

The high-quality, artefact-light image samples of FacesHQ+ also impacted the behaviour of the (admittedly under-fitted) models. All FacesHQ+ images, real or fake and regardless of provenance, possess large quantities of both high and low activation regions. In contrast with the lower-resolution images of GANFP, the FacesHQ+ optimized models as interpreted through their saliency maps appear to include higher-level features in their decision-making. For instance, highly activated regions are prevalent around human eyes, whereas background regions tend to strongly influence the activation patterns for StyleGAN images in opposing directions depending on the source. Moreover, although most CelebA-HQ images are strongly asserted to be \texttt{real}, many FFHQ images either straddle the threshold or are misrecognized as \texttt{fake}; note that only the external StarGANv2 set is derived from the CelebA-HQ real set\footnote{We assume by domain-wise similarity that the TPDNE-based GMs are trained on FFHQ. The \texttt{generated.photos} GM is trained on their private database.}. The StarGANv2 set is almost entirely dismissed as \texttt{real} with high confidence, while primary false negatives for sources included in the training set lie closer to the threshold. For source attribution, false positives rarely occur but false negatives outnumber true positives, while true positives tend to have low scores. The secondary output maps may differ drastically from primary outputs, but sometimes match each other.

Remarkably, the performance penalty of our model when fitted to augmented FacesHQ+ images is not as pronounced as in GANFP. Figure \ref{fig:localization_samples_faceshqp} depicts a set of multi-augmented images correctly detected and attributed. Prior to Phase II retraining, primary false negatives are predominant on the multi-augmented set; the more augmentations applied to a synthetic image, the more likely it would be incorrectly dismissed as \texttt{real} with high confidence. However, the augmentation-aware model inverts the bias in favour of false positives. Additive augmentations generally lead to model confusion even after retraining, but we also observed edge cases where the introduction of augmentations somehow rectified incorrect primary predictions on the same image. Compression and cropping still affect source attribution more heavily than deepfake detection, but unlike with GANFP, blurred images now correspond to slight drops in model activations. Additionally, as recorded in table \ref{tab:faceshqp_att}, almost all StarGANv2 images are not ascribed to any learned source, though augmentations also cause perplexing corrections in primary outputs and enhanced confusion in secondary outputs. Given that the primary layers for the FacesHQ+ models have apparently learned some higher-level features (whether inadvertently or otherwise), we conjecture that the secondary modules are incapacitated by the irrelevance of these features in facilitating attribution, exacerbating the destructive effects of image augmentations on attributability.

We anticipated that model performance would improve on individually augmented datasets due to the use of less tainted primary modules. This is obvious in Figure \ref{fig:localization_samples}, where the compression-aware GANFP ProGAN classifier succeeds in detection and attribution of subtly compressed images. Although the primary modules excel in deepfake detection despite image augmentations albeit at an increased false positive rate, the attached secondary modules still regularly commit binary classification errors (especially between MMDGAN and CramerGAN in GANFP) with associated erroneous activations. In light of previous observations from sections \ref{deepfake_detection} \& \ref{source_attribution}, it can be affirmed that image source attribution using GM fingerprints is not only considerably more difficult than deepfake detection, but also more susceptible to the effects of simple image perturbations, particularly in the frequency domain.

\subsection{Declaration of Deficiencies}

All our models fail to achieve satisfactory recall scores for source attribution on extensively augmented datasets. This implies that image perturbations have a greater disruptive impact on source-specific fingerprints. Enabling subsequent fine-tuning of the primary module during attribution training might resolve this issue, but would compromise the integrity and modularity of the semi-decentralized design. We acknowledge that subjecting images to successive combinations of na\"{i}ve perturbations can eventually defeat any robust classifier contingent on inherent image features; but this is relatively unlikely to occur, and may even be counterproductive for human-centred deception in certain scenarios. We did not conduct any model resilience tests using adversarial examples, though \citet{Carlini_2020} have previously managed to completely desensitize the \texttt{gandct-conv} baseline with bit-flip attacks. Similar GM hyperparameters are found to leave fingerprints similar enough to induce confusion in reactive attribution methods, though this also motivated the development of model parsing.

At higher resolutions, the pixel-input model appears to learn higher-level features that are barely appropriate for deepfake detection and practically futile for source attribution. These uncertain ``fingerprints'' also fail to generalize to synthetic images that radically differ from the training distribution, though they remain effective at disproving attribution. It remains unknown how our models would react to partially manipulated imagery that violate the binary distinction between real and fake, even after accounting for post-processing artefacts found in \texttt{real}-labelled data. Furthermore, we only considered the domain of human portraits, which are presently the most common application of synthetic imagery but contain specific features that differ from other domains.

Overall, we doubt that reactive attribution is sustainable in the long term when considering the rapid development of generative ML, techniques applied to conceal their applications, and changing perspectives on what constitutes authenticity. Our solution may improve extensibility, efficiency, and applicability; but like other fingerprint-based methods, nothing is guaranteed once the fingerprints cease to exist. Most importantly, our supervised learning method is entirely dependent on the existence of a comprehensive and diverse dataset of curated image samples, objectively and truthfully labelled beforehand according to their sources, all of which must be accessible to the custodian of the pre-trained primary module and the developers of secondary modules. This necessitates obtaining and isolating sufficient specimens from novel, undocumented sources via other means, which is beyond the scope of this study. Proactive attribution is not expected to fare any better if the cooperation of all GM developers and digital content owners is still required.

\section{Conclusion and Future Work}

We revisited the source attribution problem of GAN-generated synthetic images as a series of binary classification tasks relevant for real-world scenarios, and designed a modular neural classifier that employs transfer learning to leverage the conceptual relationship between deepfake detection and attribution for rapid deployment. We then evaluated the model's resilience against common image perturbations easily employed to defeat intrinsic fingerprint-based image forensic algorithms, and conclude whether reactive attribution remains technically feasible and sustainable for synthetic images before even considering more sophisticated attacks. We observed that binary-class attribution of image generative models (GMs) is feasible using only learned intrinsic fingerprints, but is subject to reduced sensitivity when dealing with heavily degraded images or technically similar GMs. Our model performs on par with existing benchmarks on deepfake detection, and trains more efficiently than them on source attribution under ideal conditions. Moreover, we demonstrate that implementation of class activation mapping is feasible as a means of model interpretation and validation for deepfake forensics.

Various opportunities are designated for refinement of our methods and reaffirmation of our findings. Given the restricted complexity of this initial study, the logical next step would be to verify the hypothesized scalability benefits and full potential of our proposed model in more realistic contexts. As the attribution modules are limited by the quality of features supplied by the transfer-learned primary module, we recommend replacing the latter with large-scale general-purpose classifier architectures pre-optimized for source-agnostic deepfake detection. This would further improve adaptability and external validity, while the increased development cost of the deeper primary module is negated by minimizing the need for retraining. Likewise, the effects of other hyperparameters should be investigated, including different model initialization, optimizers, secondary module topologies, and loss functions (e.g. increasing the binary threshold). Regularization losses are also preferred to suppress model self-contradictions.

Subsequent experiments are to use large-scale and appropriately augmented datasets that provide options for out-of-distribution validation and resilience tests. Other content domains such as LSUN objects and deepfake video corpora are also suited for investigation. The 100 GM dataset by \citet{ModelParsing2021} provides a representative and heterogeneous foundation for further studies; though at merely 1000 samples per source, further model simplification is desirable to prevent overfitting. Federated learning over all publicly accessible deepfake datasets may provide the missing link towards ensuring generalizable features in the decentralized paradigm. Additionally, feature localization should be expanded to accommodate the often-neglected case of \emph{partially synthetic} images. Ultimately, researchers might have to abandon supervised strategies and switch to watermarking or reverse engineering to accomplish universal attribution of synthetic media.

\printbibliography

\end{document}